\documentclass[letterpaper,journal]{IEEEtran}
\usepackage{amsmath,amsfonts}

\DeclareMathOperator*{\argmax}{argmax} 
\usepackage{hhline}
\usepackage{algorithm}
\usepackage{array}
\usepackage{xcolor}
\usepackage[caption=false,font=normalsize,labelfont=sf,textfont=sf]{subfig}
\usepackage[normalem]{ulem}
\usepackage{textcomp}
\usepackage{stfloats}
\usepackage{enumitem}
\usepackage{url}
\usepackage{verbatim}
\usepackage{graphicx}
\usepackage{arydshln}
\usepackage{cite}
\usepackage{booktabs}
\usepackage{multirow}
\usepackage{color}
\usepackage{hyperref}
\usepackage{titlesec}
\usepackage{algpseudocode}
\errorcontextlines\maxdimen

\makeatletter
    \newcommand*{\algrule}[1][\algorithmicindent]{\makebox[#1][l]{\hspace*{.5em}\thealgruleextra\vrule height \thealgruleheight depth \thealgruledepth}}%
\newcommand*{\thealgruleextra}{}
\newcommand*{\thealgruleheight}{.75\baselineskip}
\newcommand*{\thealgruledepth}{.25\baselineskip}

\newcount\ALG@printindent@tempcnta
\def\ALG@printindent{%
    \ifnum \theALG@nested>0
        \ifx\ALG@text\ALG@x@notext
        \else
            \unskip
            \addvspace{-1pt}
            \ALG@printindent@tempcnta=1
            \loop
                \algrule[\csname ALG@ind@\the\ALG@printindent@tempcnta\endcsname]%
                \advance \ALG@printindent@tempcnta 1
            \ifnum \ALG@printindent@tempcnta<\numexpr\theALG@nested+1\relax
            \repeat
        \fi
    \fi
    }%
\usepackage{etoolbox}
\patchcmd{\ALG@doentity}{\noindent\hskip\ALG@tlm}{\ALG@printindent}{}{\errmessage{failed to patch}}
\makeatother

\newbox\statebox
\newcommand{\myState}[1]{%
    \setbox\statebox=\vbox{#1}%
    \edef\thealgruleheight{\dimexpr \the\ht\statebox+1pt\relax}%
    \edef\thealgruledepth{\dimexpr \the\dp\statebox+1pt\relax}%
    \ifdim\thealgruleheight<.75\baselineskip
        \def\thealgruleheight{\dimexpr .75\baselineskip+1pt\relax}%
    \fi
    \ifdim\thealgruledepth<.25\baselineskip
        \def\thealgruledepth{\dimexpr .25\baselineskip+1pt\relax}%
    \fi
    \State #1%
    \def\thealgruleheight{\dimexpr .75\baselineskip+1pt\relax}%
    \def\thealgruledepth{\dimexpr .25\baselineskip+1pt\relax}%
}
\usepackage{etoolbox}

\begin{document}

\title{Adversarial Wear and Tear: Exploiting Natural Damage for Generating Physical-World Adversarial Examples}

\author{Samra Irshad, Seungkyu Lee, Nassir Navab, Hong Joo Lee, and Seong Tae Kim

\thanks{This work was supported in part by the Institute of Information and Communications Technology Planning and Evaluation (IITP) Grant funded by the Korea Government (MSIT) under Grant RS-2022-II220078, Grant IITP-2025-RS-2023-00258649, and by the National Research Foundation of Korea (NRF) grant funded by the Korea government (MSIT) (No. RS-2024-00334321). This work was also supported by the Global Learning \& Academic Research Institution for Master's–PhD Students and Postdocs (G-LAMP) Program of the National Research Foundation of Korea (NRF), funded by the Ministry of Education (No. RS-2025-25442355). \textit{(Corresponding authors: Seong Tae Kim and Hong Joo Lee)}}
\thanks{Samra Irshad and Seungkyu Lee are with Department of Computer Science and Engineering, Kyung Hee University, Yongin, 17104, South Korea (e-mail: samra@khu.ac.kr, seungkyu@khu.ac.kr)}
\thanks{Nassir Navab and Hong Joo Lee are with Technical University of Munich, Germany (e-mail: nassir.navab@tum.de, hongjoo.lee@tum.de). Hong Joo Lee is also affiliated with Seoul National University of Science and Technology, South Korea.}
\thanks{Seong Tae Kim is with Department of Computer Science and Engineering, Kyung Hee University and also with G-LAMP NEXUS Institute, Kyung Hee University, Yongin, 17104, Republic of Korea (e-mail: st.kim@khu.ac.kr)}}

\markboth{Journal of \LaTeX\ Class Files,~Vol.~14, No.~8, August~2021}%
{Shell \MakeLowercase{\textit{et al.}}: A Sample Article Using IEEEtran.cls for IEEE Journals}

\maketitle

\begin{abstract}
The presence of adversarial examples in the physical world poses significant challenges for deploying Deep Neural Networks (DNNs) in safety-critical applications such as autonomous driving, where even minor misclassifications can lead to catastrophic consequences. Most existing methods for crafting physical-world adversarial examples are ad-hoc and deliberately designed, relying on temporary modifications like shadows, laser beams, or stickers that are tailored to specific scenarios. In this paper, we introduce a new class of physical-world adversarial examples, \textit{AdvWT}, which draws inspiration from the naturally occurring phenomenon of `wear and tear', an inherent property of physical objects. Unlike manually crafted perturbations, `wear and tear' emerges organically over time due to environmental factors, as seen in the gradual deterioration of outdoor signboards. To achieve this, \textit{AdvWT} follows a two-step approach. First, a GAN-based, unsupervised image-to-image translation network is employed to model these naturally occurring damages, particularly in the context of outdoor signboards. The translation network encodes the characteristics of damaged signs into a latent `damage style code'. In the second step, we introduce adversarial perturbations into the style code, strategically optimizing its transformation process. This manipulation subtly alters the damage style representation, guiding the network to generate adversarial images where the appearance of damages remains perceptually realistic, while simultaneously ensuring their effectiveness in misleading neural networks. Through comprehensive experiments on two traffic sign datasets, we show that \textit{AdvWT} effectively misleads DNNs in both digital and physical domains. \textit{AdvWT} achieves an effective attack success rate, transferability, robustness, and a more natural appearance compared to existing physical-world adversarial examples. Additionally, integrating \textit{AdvWT} into training enhances a model's generalizability to real-world damaged signs. We have released the code for \textit{AdvWT} at \url{https://github.com/samra-irshad/AdvWT}.
\end{abstract}

\begin{IEEEkeywords}
Adversarial examples, Image-to-image translation, Robustness.
\end{IEEEkeywords}

\section{Introduction}
\IEEEPARstart{I}{n} recent years, Deep Neural Networks (DNNs) have become increasingly popular for a wide range of computer vision tasks. Despite their success, DNNs remain vulnerable to adversarial perturbations, which significantly impact their reliability. The susceptibility of DNNs to adversarial threats poses significant concerns for their deployment in safety-critical applications, such as autonomous driving and healthcare \cite{Bojarski2016, Fu2022, Mnih2015}. For instance, an adversary could manipulate an outdoor traffic sign in the real world, potentially causing vehicles to make dangerous decisions \cite{Kong2019}. Recently, physical-world adversarial attacks have gained substantial attention due to the severe risks of catastrophic consequences they present \cite{Wang2023, Duan_2021_CVPR, Zhong_2022_CVPR}.

Classified by the manner in which perturbation patterns are realized, two principal methods have emerged for crafting adversarial examples in the physical world. The first involves introducing perturbations via optical patterns, such as laser beams \cite{Duan_2021_CVPR}, shadows \cite{Zhong_2022_CVPR}, or projected light \cite{gnanasambandam2021optical}. The second approach entails directly altering the target object, by applying meticulously crafted adversarial patches \cite{brown2017adversarial, Eykholt2018} or synthetic styles \cite{Duan2020}. While these approaches can be effective, their applicability depends on specific conditions. Optical pattern-based attacks typically require controlled illumination or low-light environments \cite{Zhong_2022_CVPR, Duan_2021_CVPR}, while adversarial patches generally do not prioritize stealth and remain visually conspicuous \cite{brown2017adversarial}. 

Additionally, to ensure real-world robustness, existing physical-world attacks inject larger perturbations to withstand environmental variations, thereby compromising stealth and leading to visually unnatural artifacts. Moreover, most of the existing physical-world adversarial perturbations are crafted with a single specific pattern (e.g., shadows \cite{Zhong_2022_CVPR}, patches \cite{Eykholt2018}). Utilizing a uniform pattern enables defense mechanisms to identify the distribution of adversarial patterns, facilitating effective countermeasures against such attacks \cite{wang2022shadows} \cite{xiang2021patchguard} \cite{Samangouei2018}. 

Building on the observation that objects in outdoor environments are subject to natural degradation over time, which can alter their appearance and potentially mislead DNNs, we introduce \textbf{Adversarial Wear and Tear} (\textit{AdvWT}). Unlike existing attacks, \textit{AdvWT} perturbations emerge passively over time, making them both persistent and more difficult to mitigate. Fig. \ref{fig:fig1} compares real-world damaged signs with \textit{AdvWT}-generated adversarial examples, demonstrating how our proposed damage simulation seamlessly integrates realistic degradation patterns.

\textit{AdvWT} introduces a new paradigm by leveraging real-world degradation as a stealthy adversarial cue. Unlike existing physical adversarial examples, which can be reversed by removing the applied perturbation, \textit{AdvWT} introduces modifications that are difficult to remove without physically restoring or replacing the object. We simulate \textbf{wear and tear} by learning a damage representation that captures degradation patterns. Using an unsupervised GAN-based framework \cite{Choi2019}, we learn and optimize a latent damage representation to introduce adversarial perturbations that mislead DNNs while maintaining visual realism. We particularly focus on outdoor traffic signs, a safety-critical component of autonomous driving systems where recognition errors can have severe consequences. These signs naturally degrade through weathering, fading, corrosion, dirt accumulation, and structural wear, making them a realistic and high-impact testbed for our attack. \footnote{See our discussion on the definition of \textit{natural damage} in the Section \ref{sec:damageness}.}

 Based on the characteristics of our proposed adversarial examples, we identify four key properties for comparison with existing physical-world attacks: \textbf{Gradual Evolution}, \textbf{Occurrence Rate}, \textbf{Pattern Diversity}, and \textbf{Persistency} (Fig. \ref{fig:fig2}). \textbf{Gradual Evolution} refers to damage that accumulates over time from environmental exposure, unlike instantaneous light- or projection-based attacks. \textbf{Occurrence Rate} describes how often a perturbation occurs in practice, with natural sign damage far more prevalent than laser or shadow attacks. \textbf{Pattern Diversity} captures the range of visual variations, as \textit{wear and tear} can manifest in many forms, such as cracks, fading, and discoloration. \textbf{Persistency} describes the durability of perturbations, since physical degradation persists until repair, unlike transient light or shadow effects. The main contributions of this work are as follows:
\begin{figure*}[t]
    \centering
    \setlength{\belowcaptionskip}{-7pt}
        \includegraphics[width=0.75\linewidth]{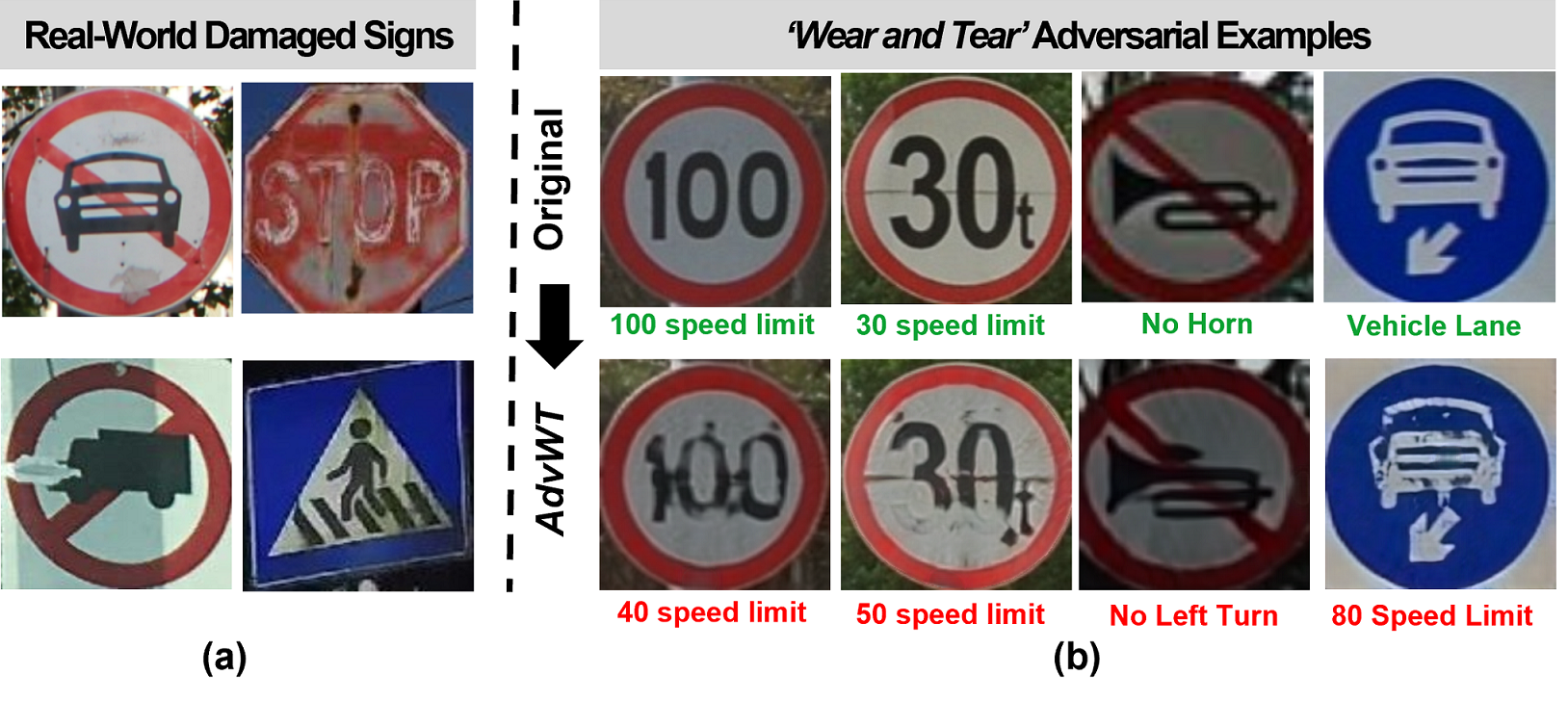}
    \caption{Adversarial Wear \& Tear Examples. (a) Damaged traffic signs observed in real-world. (b) Original traffic Signs (first row) with correct predictions in green font and \textit{adversarially} damaged traffic signs generated by \textbf{\textit{AdvWT}} (second row) with misclassified labels in red font. Our proposed adversarial perturbations not only resemble real-world degradation but also successfully manipulate model predictions.}
    \label{fig:fig1}
\end{figure*}
\begin{itemize}[leftmargin=*]
\item We propose \textit{AdvWT}, an attack that exploits the natural \textit{wear and tear} of physical objects to craft stealthy adversarial examples. It produces realistic damage on traffic signs, with perturbations that appear as cracks, corrosion, peeling paint, or dirt accumulation. 
\item We introduce a framework for simulating naturally damaged traffic signs that captures the stochastic and progressive nature of real-world degradation, and use it to evaluate the impact of \textit{adversarial damage} on recognition models.
\item Our approach for damage simulation utilizes a GAN-based unsupervised Image-to-Image (I2I) translation framework \cite{Choi2019} to learn the representations of damaged traffic signs. Subsequently, we explore the latent space to identify adversarially damaged codes, enabling a controlled generation of realistically degraded signs.
\item Through extensive experiments, we verify that the proposed \textit{AdvWT} successfully fools DNNs in digital and physical domains. The adversarial perturbations generated by \textit{AdvWT} demonstrate effective attack success rate, transferability, robustness, naturalness, and generalizability against real-world damaged road signs.
\end{itemize}
\section{Related Work}
\subsection{Adversarial Examples}
First introduced by \cite{Szegedy2013}, adversarial examples are subtle and imperceptible perturbations that can cause DNNs to misclassify when added to an input. Since their discovery, the robustness of DNNs against various adversarial threats has been extensively studied \cite{Carlini2016, su2019one, Dong_2018_CVPR, shen2022}. Adversarial attacks are commonly classified by (a) Adversary’s knowledge of the target model (black- or white-box), (b) Deployment scenario (digital or physical), (c) Attack objective (targeted or untargeted), and (d) Attack mechanism (contact-based or contactless).
\subsubsection{White-Box and Black-Box Attack Settings}
Based on the adversary's knowledge of the target model, adversarial attacks are categorized as white-box or black-box \cite{akhtar2018threat}. White-box attacks assume partial or full access to the model (e.g., its architecture and gradients), whereas black-box attacks rely on surrogate models without direct access to the model.
\subsubsection{Targeted or Untargeted Attacks}
Adversarial attacks can be formulated with either untargeted or targeted objectives \cite{akhtar2018threat}. Untargeted attacks induce any misclassification, whereas targeted attacks force the prediction of a specific class.
\subsubsection{Digital and Physical attack scenarios}
Adversarial attacks exist in both digital and physical domains. Digital attacks typically constrain perturbations using norms such as $l_{\infty}$, $l_{2}$, or $l_{0}$ to ensure imperceptibility \cite{Goodfellow2014,Carlini2016}. In contrast, physical-world attacks are often more unconstrained and conspicuous, as they must remain effective under real-world transformations such as lighting, pose, and distance \cite{Sharif2016,Ziwei2025}. Given the ubiquity of such adversarial vulnerabilities in real-world environments, numerous studies have investigated their implications.
\subsection{Physical-world Adversarial Examples}
The feasibility of adversarial examples in the physical world was first demonstrated by \cite{Kurakin2016}. These adversarial examples are typically designed to be printed and subsequently captured by a camera \cite{Kurakin2016}. Existing physical-world attacks can be categorized by their mechanism as contactless or contact-based adversarial examples.
\subsubsection{Contactless Physical Adversarial Examples}
Contactless physical adversarial attacks manipulate optical phenomena, such as illumination, without physically altering the object. Early work explored camera- and lighting-based effects, including roller shutter manipulation \cite{Sayles2021} and reflection-based perturbations \cite{yunfei2020}. Subsequent studies expanded this direction by using specialized light sources and illumination patterns \cite{Sayles2020, gnanasambandam2021optical}. Duan et al. \cite{Duan_2021_CVPR} demonstrated that optimized laser projections can reliably mislead DNNs, while Zhong et al. \cite{Zhong_2022_CVPR} proposed shadow-based attacks using simulated shadows. More recently, Wang et al. \cite{Wang2023} introduced reflected light attacks, and Hu et al. \cite{Hu2024} explored neon beam-based optical perturbations. Although highly stealthy, these methods often depend on specific lighting and environmental conditions.
\begin{figure}[t]
    \centering
        \includegraphics[width=0.99\linewidth]{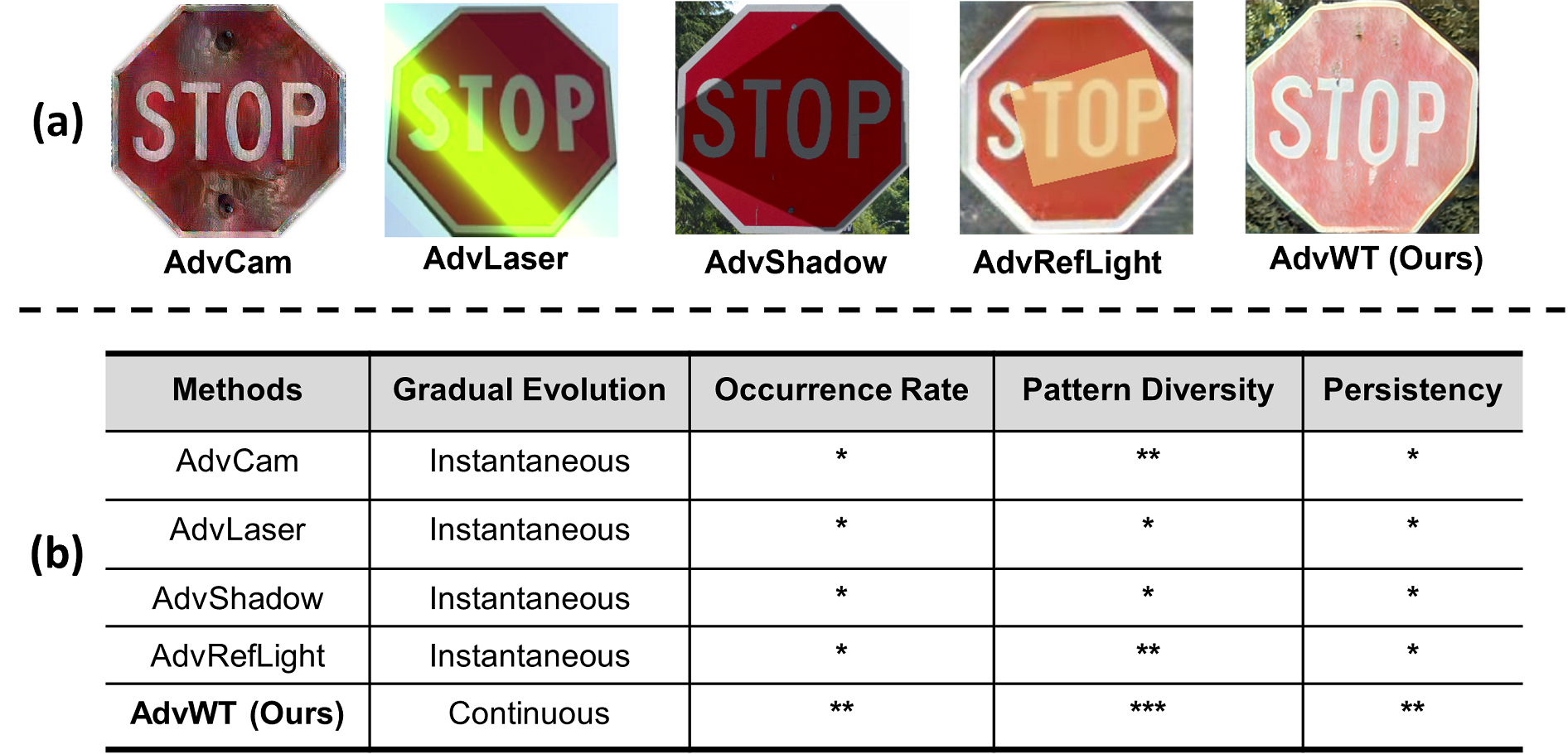}
    \caption{Comparison of physical-world adversarial examples: (a) Examples of AdvCam \cite{Duan2020}, AdvLaser \cite{Duan_2021_CVPR}, AdvShadow \cite{Zhong_2022_CVPR}, RFLA \cite{Wang2023}, and \textit{AdvWT} (ours). (b) Methods are compared based on gradual evolution, occurrence rate, pattern diversity, and persistency. Asterisks indicate presence: * (low), ** (moderate), *** (high)}
    \label{fig:fig2}
\end{figure}
\subsubsection{Contact-based Physical Adversarial Examples}
Contact-based physical adversarial attacks alter an object’s appearance through direct physical interaction, such as adversarial patches or stickers \cite{Sharif2016,brown2017adversarial,Eykholt2018}, as well as naturally occurring surface alterations like scratches \cite{Loris2023}. Style-based contact attacks further modify appearance via neural style transfer or latent-space perturbations \cite{Duan2020,song2018constructing,fu2023styleadv}. While effective, these approaches often introduce artificial textures or colors due to globally applied style statistics, which can distort object semantics \cite{huang2017arbitrary,li2017demystifying}.

In contrast, our approach is grounded in natural degradation processes observed on outdoor traffic signs, such as cracks, rust, peeling paint, and dirt. We model a damage space from paired clean and naturally damaged traffic signs using a GAN-based image-to-image translation framework with cycle-consistency constraints \cite{zhu2017unpaired,choi2020stargan}, enabling structured and domain-aware perturbations that reflect realistic wear and tear. Our work is also related to ACA \cite{chen2023content}, which generates unrestricted style variations; however, our method explicitly captures the evolving, stochastic, and persistent nature of real-world damage, resulting in semantically meaningful degradations tailored to traffic sign recognition systems.
\subsection{Adversarial Defense Methods}
Adversarial defense methods seek to improve robustness against adversarial perturbations through various strategies \cite{Goodfellow2014}. Adversarial training (AT) is one of the earliest and most influential defenses, incorporating adversarial examples during training to promote robust feature learning \cite{Madry2017}. Beyond AT, input purification methods have been proposed to remove adversarial noise at test time, including diffusion-based adversarial purification \cite{nie2022diffusion} and DISCO, which reconstructs clean images via localized implicit functions \cite{ho2022disco}. However, most prior physical-world attacks, such as adversarial laser \cite{Duan_2021_CVPR}, shadow \cite{Zhong_2022_CVPR}, and neon light attacks \cite{Hu2024}, primarily rely on adversarial training as their defense baseline.
\section{Proposed Method}\label{sec:prop}
\subsection{Preliminaries}
Let $\mathcal{F}(x): \mathbb{R}^{h} \rightarrow \mathbb{R}^{q}$ denote a DNN-based image classifier that outputs class scores for an input image $x$ with ground-truth label $y$ where $h$ is the input dimensionality and $q$ is the number of classes. To classify $x$, the model produces probability scores using \textbf{softmax} function.
\begin{equation}\label{eq:eq12}
p(g|x) = \dfrac{e^{\phi_g(x)}}{\sum_{j=1}^{q} e^{\phi_j(x)}}
\end{equation}
where $p(g|x)$ represents the probability of $x$ belonging to class $g$. $\phi_{g}(x)$ is the model’s logit score for class $g$. The denominator sums over all $q$ class logits $\phi_{j}(x)$, ensuring the probabilities sum to 1. The final predicted label $\hat{y}$ is determined by selecting the class with the highest probability.
\begin{equation}\label{eq:eq13}
\hat{y} = \arg\max_{g \in {1, \dots, q}} p(g|x)
\end{equation}
 The objective of an adversarial attack is to induce a misclassification such that $\arg\max_{g} p(g|x),  \neq  \arg\max_{g} p(g|x_{adv})$ where $g \in \{1, \dots, q\}$. The perturbation applied to generate $x_{adv}$ must remain subtle and visually natural to human observers. In the physical world, adversarial examples are subject to additional constraints beyond visual subtlety. Physical-world adversarial perturbations should preserve photorealism, physical plausibility, and long-term persistence in real-world conditions.
 \subsection{Adversarial Wear and Tear Examples}
 Our proposed method, \textit{AdvWT}, leverages visual effects of physical wear and tear to generate adversarial examples. We formulate this task as a \textbf{damage-aware style learning} problem, aiming to model appearance variations between clean and damaged traffic signs. Specifically, we employ StarGAN-v2 \cite{Liu2021} to learn latent style codes that capture diverse real-world degradation patterns. We define two stylistic domains: \textbf{(a) a damaged domain}, representing wear and tear, and \textbf{(b) a clean domain}, corresponding to undamaged signs. The learned damage style code encodes structural and textural degradation while largely preserving sign identity, enabling the synthesis of visually coherent damaged variants that serve as a basis for subsequent adversarial manipulation in latent space.
\subsubsection{Simulating Damaged Traffic Signs}\label{sec:mod}
\begin{figure*}[t]
    \centering
        \includegraphics[width=0.88\linewidth]{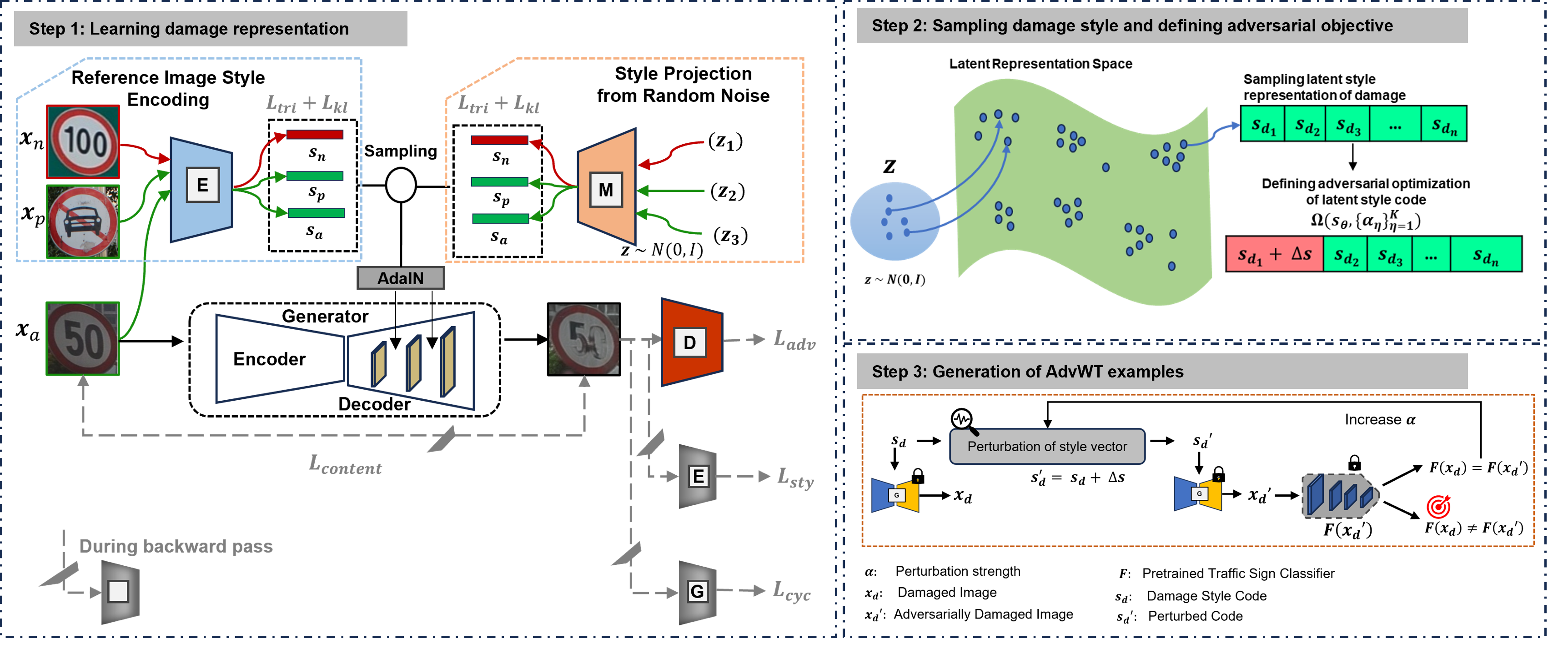}
\caption{Our approach for generating \textit{AdvWT} examples. Step 1: Train an image-to-image translation model to represent both clean and damaged traffic signs, learning a latent space of realistic `wear and tear'. Step 2: Sample latent noise $z$ from a Gaussian and map it via the style projector $M(z)$ to obtain damage styles. Step 3: An adversarial style search iteratively perturbs the damage style code $s_d$ with $\Delta s$, generating adversarial examples whose perturbation strength $\alpha$ is tuned to maximize misclassification by the target classifier $\mathcal{F}$.}
    \label{fig:fig3}
\end{figure*}

We employ StarGAN-v2 to learn a style-based latent representation of traffic sign damage, enabling the photorealistic simulation of degradation patterns—an underexplored direction in adversarial robustness. As illustrated in Fig. \ref{fig:fig3}, the framework comprises a \textbf{Style Encoder} $E(x)$, a \textbf{Noise-to-Style Mapping Network \( M(z) \)}, an image \textbf{Generator \( G(x; s) \)}, and a \textbf{Discriminator \( D \)}. Style vectors corresponding to clean and damaged signs are denoted as \( s_c \) and \( s_d \), respectively and are obtained either from reference images via \( E(x) \) or from random noise \( z \sim N(0,1) \) through \( M(z) \). The generator integrates style information using Adaptive Instance Normalization (AdaIN) \cite{huang2017arbitrary,karras2019style}. We provide detailed descriptions of each architectural module in Appendix A of the Supplementary Materials. To promote high-fidelity translations that preserve structural integrity while introducing realistic damage, we optimize a weighted combination of seven objectives \cite{Choi2019,Liu2021}, detailed in Eq. \ref{eq:eq1}. 
\begin{equation}\label{eq:eq1}
  \lambda_{1} L_{sty} +   \lambda_{2}L_{ds} +   \lambda_{3}L_{cyc} +   \lambda_{4}L_{adv} +   \lambda_{5}L_{tri} +   \lambda_{6}L_{kl} +   \lambda_{7}L_{cont}
 \end{equation}
\( L_{sty} \) is the \textit{style reconstruction loss} \cite{huang2018multimodal} which ensures that the generated image \( E(G(x, s)) \) accurately reflects the desired `wear and tear' characteristics encoded in $s$, such as dirt, fading, or scratches. By aligning the generated image with the target damage style, $L_{sty}$ promotes realistic and domain-consistent degradation. 
  \begin{equation}
    L_{sty} = \mathop{{}\mathbb{E}_{x\sim X, s\sim S}[||\textbf{s}-E(G(\textbf{x},\textbf{s}))||_{1}]}
 \end{equation}
 \( L_{ds} \) is the \textit{diversity sensitivity loss} \cite{Choi2019}, whereas $s_{1}$ and $s_{2}$ are style codes sampled from the same domain. This loss prevents mode collapse and encourages diverse degradation patterns.
   \begin{equation}
    L_{ds} = -\mathop{{}\mathbb{E}_{x\sim X_{i}, (s_{1},s_{2})\sim S_{j}}[||G(\textbf{x},\textbf{s}_{1}-G(x,s_{2})||_{1}]}
 \end{equation}
 \( L_{cyc} \) is the \textit{cyclic consistency loss} \cite{Choi2019}, that pushes the generator $G$ to preserve the semantic identity of the input sign $x$ to remain unchanged during translation. This is crucial for maintaining the functional meaning of the sign while applying realistic damage patterns.
  \begin{equation}
    L_{cyc} = \mathop{{}\mathbb{E}_{x\sim X, s\sim S}[||x-G(G(\textbf{x},\textbf{s}),E(x)||_{1}]}
 \end{equation}
\( L_{adv} \) is the \textit{adversarial loss} \cite{Choi2019}, that predicts if the image $x$ is a real image of its domain or a fake image. 
   \begin{equation}
    L_{adv} = \mathop{{}\mathbb{E}_{x\sim X_{i},s\sim S_{j}}[||\text{log}D_{i}(\textbf{x})+\text{log}(1-D_{j}(G(\textbf{x},\textbf{s})))]}
 \end{equation}
\( L_{tri} \) enforces domain disentanglement by clustering style codes from the same domain while separating those from different domains.
Following the standard formulation \cite{schroff2015facenet}, an anchor style code \( s_a \) and a positive style code \( s_p \) are both extracted from the same domain (\( s_a, s_p \in S_i \)), whereas the negative style code \( s_n \) is drawn from a different domain (\( s_n \in S_j, j \neq i \)), as shown in Fig. \ref{fig:fig3}. These style representations are obtained by sampling real images and encoding them through the style encoder. A margin parameter \( \beta \) (set as 0.1 in our case) controls inter-domain separation, encouraging a well-structured latent space that preserves domain-specific damage characteristics.
This loss helps the model learn domain-specific damage patterns more effectively by maintaining a well-separated and interpretable latent style space.
 \begin{equation}
    L_{tri} = \mathop{{}\mathbb{E}_{(\textbf{s}_{a},\textbf{s}_{p},\textbf{s}_{n})\sim\textbf{S}}}[\text{max}(||\textbf{s}_{a}-\textbf{s}_{p}||-||\textbf{s}_{a}-\textbf{s}_{n}||+\beta,0)]
 \end{equation}
\( L_{kl} \) is the KL-divergence loss that enforces a prior Gaussian distribution on the set of all the style codes to encourage a compact space $\textbf{S}$. Enforcing this constraint allows smooth transitions between different styles of degradation and ensures that randomly sampled latent codes correspond to realistic and semantically meaningful damage patterns.
  \begin{equation}
    L_{kl} = \mathop{{}\mathbb{E}_{s\sim S}[||D_{KL}(p(\textbf{s})||N(\textbf{0},\textbf{I}))]}
 \end{equation}
\( L_{cont} \) is the \textit{content preservation loss} \cite{Choi2019}, that constraints the generator $G$ to preserve the content of source image $x$. This loss is crucial to avoid altering the sign’s intended meaning while simulating natural degradation. In Eq. \ref{eq:eq1}, we set the values of $\lambda$ to 1, except for $\lambda_{1}$ which is set to 2. 
\begin{equation}
    L_{cont} = \mathop{{}\mathbb{E}_{x\sim X, s\sim S}[\psi (\textbf{x},G(\textbf{x},\textbf{s})))]}
 \end{equation}
\subsubsection{Damage-Aware Latent Style Optimization} 
After training the image-to-image bidirectional translation GAN, we can obtain style vectors that capture the distinct visual characteristics of clean and damaged traffic signs. The style code, an $N$-dimensional latent representation, encodes the transformation dynamics between these two domains. Given a clean traffic sign \( x_c \) and a damage style code \( s_{d} \), the generator can synthesize a realistically degraded version \( x_{d} \) at test time. To generate our proposed \textit{AdvWT} perturbations, we introduce damage latent optimization process \( \Omega \) that iteratively manipulates learned style vectors to craft adversarial examples disguised as \textit{natural physical damage}. We formulate this process as:
\begin{equation}
     \Omega(s_{\theta}, \{\alpha_{\eta}\}_{\eta=1}^{K}) =\arg\min_{s_{\theta}}\; p(\hat{y}|G(x',s_{\theta}+\alpha|s_{\theta}|)) 
     \label{eq:eqmul}
\end{equation}
The goal of the optimization process \(\Omega\) is to iteratively refine damage representation \(s_{\theta}\) such that the generated image \( G(x', s_{\theta} + \alpha |s_{\theta}|) \) minimizes the classifier’s confidence in the correct class. The details of our algorithm for finding adversarial code are provided in Algorithm~\ref{alg:alg1}. \\
\textbf{Stage I:} Using the trained style projector $M(z)$, we sample a latent noise vector $z \sim \mathcal{N}(0,1)$ and generate the initial damage style code $s_d$, which is used
to initialize the optimization process such that $s_{\theta} = s_d$.\\
\textbf{Stage II:} Using the initial $s_{\theta}$, we generate the damaged traffic sign $x_d$ and
classify it with the pre-trained model $\mathcal{F}$. Assuming the predicted
class of traffic sign $\hat{y}$ is correct, we record the confidence score
$p(\hat{y} \mid x_d)$ along with the corresponding style code $s_{\theta}$.\\
\textbf{Stage III:} We add a perturbation $\alpha$ to the style code $s_{\theta}$, applied as $\alpha |s_{\theta}|$ to modulate the extent of modification in the generated image. The perturbation strength $\alpha$ is increased from $c_{\min}=0.1$ to $c_{\max}=1.5$ over $K=15$ steps.\\
\textbf{Stage IV:} At each step $\eta$, we scale the perturbation strength $\alpha$ and sample a batch
of $T=30$ candidate style codes within the dynamically updated bounds $[s_{\theta}^-, s_{\theta}^+]$ using a Monte Carlo sampling approach. Each candidate is used to generate an adversarial image
$G(x', s_{\theta}+\alpha|s_{\theta}|)$, which is evaluated by the classifier. The perturbation that most reduces confidence is selected for refinement. The perturbation strength $\alpha$ is increased as needed until an adversarial style code is found or the maximum threshold is reached.
\begin{algorithm}[t]
\small
\caption{Pseudocode for finding adversarial damage style code}\label{alg:alg1}
\begin{algorithmic}[1]
\Statex \hspace{-\algorithmicindent}{\textsc{\textbf{Input:}}} \parbox[t]{\dimexpr 0.92\linewidth-\algorithmicindent}{Clean input $x$; Trained style projector $M(z)$ Random Noise $z$; Perturbation range $[c_{\min}, c_{\max}]$; Step size $\eta$; Number of style vectors sampled for each step $T$; Target traffic sign classifier $\mathcal{F}$}; 
\Statex \hspace{-\algorithmicindent} {\textsc{\textbf{Output: }}}Adversarial damage style code $s_{adv}$
\setcounter{ALG@line}{0}
\State $z \sim N(0,1)$\;
\State \textbf{Initialize} $s_{\theta} \gets M(z)$\;
\State Generate damaged traffic sign $x_{d} \gets G(x; s_{\theta})$
\State Predict the class label $\hat{y} = \arg\max_{g} p(g|x_d)$
\State Set $\text{Conf}_{\theta} \gets p(\hat{y} | x_d)$
\State Set $s_{\text{best}} \gets s_{\theta}$
\For{$\alpha \gets c_{\min}$ \textbf{to} $c_{\max}$ \textbf{with step} $\eta$}
\State Compute perturbation: $\Delta s_{\theta} \gets \alpha |s_{\theta}|$
\State Define bounds: $s_{\theta}^{+} \gets s_{\theta} + \Delta s_{\theta}, \quad s_{\theta}^{-} \gets s_{\theta} - \Delta s_{\theta}$
\State \textit{// Generate and evaluate $T$ images}
\State From $[s_{\theta}^{-}, s_{\theta}^{+}]$ range, sample $T$ style vectors
\State Generate images: $\{x'_i\}_{i=1}^{T} \gets \{G(x; s_{\theta_i})\}_{i=1}^{T}$
 \State Compute classifier scores: $\{p_i\}_{i=1}^{T} \gets \{\mathcal{F}(x'_i)\}_{i=1}^{T}$
 \State \textit{// Check for adversarial success}
     \For{$i = 1$ to $T$}
        \If{$\argmax p_i \neq \argmax p(x_d)$} 
            \State \textbf{Return} $s_{\text{adv}} = s_{\theta_i}$ 
        \ElsIf{$p_i < \text{Conf}_{\theta}$}
        \State \textit{// Track most adversarial-like style vector}
            \State $\text{Conf}_{\theta} \gets p_i$
            \State $s_{\text{best}} \gets s_{\theta_i}$
        \EndIf
    \EndFor
       \If{No adversarial style code found}
        \State $s_{\theta} \gets s_{\text{best}}$ \Comment{Update style vector for next iteration}
        \State Increase perturbation factor: $\alpha \gets \alpha + \eta$
        \State Expand search bounds: $\Delta s_{\theta} \gets \alpha |s_{\theta}|$
        \State \algorithmicdo \quad Repeat search with updated parameters
    \EndIf
\EndFor
\end{algorithmic}
\label{alg1}
\end{algorithm}
\section{Experiments}
Here we describe the datasets and experiments conducted to evaluate \textit{AdvWT}. Our framework assesses effectiveness across digital and physical settings, black-box transferability, perceptual naturalness, adversarial robustness, stability under weather changes, and out-of-distribution generalization, along with an analysis of the induced perturbation patterns.
\subsection{Datasets}
We use a dataset of 4,529 traffic sign images (2,272 damaged and 2,257 undamaged), collected from online sources and public benchmarks (details are provided in Appendix B of the Supplementary Materials). This dataset is used to train the image-to-image translation model, while two additional datasets are used for training traffic sign recognition models:\\
\textbf{(i). Hybrid Dataset:} This dataset comprises 2,199 training images and 604 test images, spanning 14 road sign classes.\\
\textbf{(ii). GTSRB-HQ Dataset:} Constructed by filtering high-quality images from the GTSRB traffic dataset, it contains 1,504 training images and 309 test images, covering 10 road sign classes.
\subsection{Threat Model}
We consider an untargeted adversarial attack, where the goal is to perturb input images so the classifier outputs any incorrect label rather than a specific one. This setting reflects realistic conditions: natural or induced degradations (e.g., fading, occlusion, surface damage) rarely enforce a precise misclassification but consistently reduce model robustness. In particular, we focus on a score-based gray-box threat model in which the adversary uses the same model for crafting and evaluation, but only observes its softmax confidence scores \cite{Zhong_2022_CVPR}. Model parameters, architecture, and gradients remain hidden. Perturbations are iteratively updated based on the classifier's output scores, and optimization continues until a misclassification is achieved.
\subsection{Implementation Details}
We train the image-to-image translation model (Section \ref{sec:prop}) in an unsupervised manner using clean and damaged traffic sign images (resized to 256 $\times$ 256) with only set-level labels (clean vs. damaged). We have released the dataset publicly\footnote{\url{https://github.com/samra-irshad/AdvWT}}. The translation model is trained for 100K iterations with a batch size of 8 on a single RTX 3090 GPU using a PyTorch implementation of StarGAN-v2 \cite{choi2020stargan} and the Adam optimizer ($\beta_{1}=0$, $\beta_{2}=0.99$), with learning rates of $10^{-4}$ for the generator, discriminator, and style encoder, and $10^{-6}$ for the mapping network. For evaluation, we use exponential moving averages of model parameters \cite{karras2017progressive}. We initialize weights with He initialization \cite{he2015delving}. We provide training loss curves (Fig. S1, S2, and S3) and t-SNE visualizations (Fig. S4) in Appendix C of Supplementary Materials.

To quantitatively evaluate the effectiveness of our proposed method, \textit{AdvWT}, we measure its attack performance against a diverse set of road sign classifiers including five CNN-based architectures; ResNet-18 (RN18) \cite{he2016deep}, ResNet-50 (RN50) \cite{he2016deep}, DenseNet-121 (DN121) \cite{huang2017densely}, EfficientNet-B7 (EffNet-B7) \cite{tan2019efficientnet}, and MobileNet (MobNet) \cite{howard2017mobilenets} as well as three transformer-based architectures; Vision Transformer (ViT-B) \cite{dosovitskiy2020image}, Swin Transformer (Swin-B) \cite{liu2021swin}, and Multi-axis Vision Transformer (MaxViT-T) \cite{tu2022maxvit} classifiers trained individually on Hybrid and GTSRB-HQ datasets. We use cross-entropy loss, Adam optimization, a batch size of 16, and a learning rate of 0.001 to train the classifiers. 
\subsection{Attack Effectiveness}\label{sec:digital_exp}
Table \ref{tab:table1} compares the attack effectiveness performance \textit{AdvWT} in terms of ASR with the other physical-world adversarial attacks\footnote{We exclude AdvCam since it is architecture-specific (VGG-19 with neural style transfer layers) and requires a target style image for each, without a clear selection protocol, making it infeasible to compare fairly in our setting.} (AdvLaser \cite{Duan_2021_CVPR}, AdvShadow \cite{Zhong_2022_CVPR}, and RFLA \cite{Wang2023}). \textit{AdvWT} consistently achieves a higher Average Attack Success Rate (Avg. ASR) across different datasets and models. In particular, \textit{AdvWT} reaches near-perfect ASR on lightweight CNNs such as ResNet-18 and MobileNet, while also maintaining strong ASR on transformer-based models. We attribute this effectiveness to the use of naturalistic \textit{wear-and-tear} perturbations that mimic realistic degradation (e.g., cracks, fading, dirt), rather than artificial overlays such as patches or shadows. By embedding adversarial cues within plausible damage, \textit{AdvWT} produces visually coherent perturbations that transfer robustly across architectures and datasets.
\begin{table*}[h]
\caption{Comparison of Attack Success Rates (ASR) (\%) for \textit{AdvWT} and other attacks}
\centering
\scriptsize
\begin{tabular}{llccccccccc}
\toprule
\multirow{2}{*}{\textbf{Datasets}} & \multirow{2}{*}{\textbf{Attacks}} & \multicolumn{5}{c}{\textbf{ASR} (CNN-based Models)} & \multicolumn{3}{c}{\textbf{ASR} (ViT-based Models)} & \multirow{2}{*}{\textbf{Avg. ASR}} \\ \cmidrule(lr){3-7} \cmidrule(lr){8-10}
& & RN18 & RN50 & DN121 & EffNet-b7 & MobNet & ViT-B & Swin-B & MaxViT & \\
\midrule
\multirow{5}{*}{Hybrid} & Benign & 3.48 & 2.15 & 13.25 & 9.27 & 5.46 & 8.61 & 17.38 & 17.38 & 15.23 \\ \cmidrule(lr){2-11}
 & AdvLaser & 55.90&	59.11&	82.28	&87.42&	78.81	&54.47	&76.16	&66.23	&70.05 \\
& AdvShadow & 73.70	&70.86	&89.57	&\textbf{94.37}	&87.58	&82.28	&62.41	&90.40	&81.40
 \\
& RFLA & 90.89&	83.61&	\textbf{95.86}	&\textbf{94.37}	&85.10&	71.19&	81.13&	89.90&	86.51 \\ \cmidrule(lr){2-11}
& \textbf{AdvWT (Ours)} & \textbf{96.2}	&\textbf{85.92}	&95.03	&93.87	&\textbf{91.04}	&\textbf{86.30}	&\textbf{90.73}	&\textbf{92.22}&	\textbf{91.41}
 \\
\cmidrule(lr){1-11}
\multirow{5}{*}{GTSRB-HQ} & Benign & 0.33	&1.62	&4.53	&3.56	& 0.0	&	0.33& 13.27	&7.12	&3.845 \\ \cmidrule(lr){2-11}
 & AdvLaser & 21.00&	66.02&	72.17&	78.96&	71.84&	41.75&	69.58&	56.63&	59.74
 \\
& AdvShadow & 52.80&	85.44&	71.20&	74.43&	73.14&	61.17&	93.53&	76.05&	73.47
 \\
& RFLA & 87.38&	89.64	&92.88	&90.29	&90.61	&83.50&	70.87&	77.67&	85.36
 \\\cmidrule(lr){2-11}
& \textbf{AdvWT (Ours)} & \textbf{99.68}	&\textbf{93.53}	&\textbf{94.50}	&\textbf{93.51}	&\textbf{94.82}&	\textbf{83.80}&	\textbf{93.85}&	\textbf{92.88}&	\textbf{93.32}
 \\
 \bottomrule
\end{tabular}
\label{tab:table1}
\end{table*}
\subsection{Transferrability of AdvWT examples}\label{sec:transfer_exp}
In real-world deployments, adversarial examples that transfer across model architectures pose a greater security risk \cite{paper2016}. To evaluate the persistence of \textit{AdvWT} perturbations on unseen models with unknown architectures, we perform black-box transferability experiments. We implement the attack across multiple surrogate architectures, including CNNs and Vision Transformers, and evaluate its transferability to various black-box targets. Transferability is measured by the targets’ Attack Success Rate (ASR). As shown in Tables \ref{tab:table2} and \ref{tab:table3}, \textit{AdvWT} attains the highest average transferability across most architectures and surrogates on both datasets, outperforming prior attacks.
These results show that \textit{AdvWT} captures cross-architecture vulnerabilities more effectively than competing methods, achieving strong black-box transfer by consistently degrading the accuracy of unseen models, underscoring its practical relevance for real-world traffic sign systems.
\begin{table*}[t]
\centering
\setlength{\tabcolsep}{5.2pt}
\caption{Evaluation of \textit{AdvWT} perturbation transferability across various black-box victim models on Hybrid dataset}\label{tab:table2}
\scriptsize
\begin{tabular}{llcccccccc@{\hspace{6pt}}c}
\toprule
\multirow{2}{*}{Surrogate} & \multirow{2}{*}{Attack} &
\multicolumn{8}{c}{Black-box Target Models} & \multirow{2}{*}{Avg. ASR} \\
\cmidrule(lr){3-10}
 & &RN18 & RN50 & DN121 & EffNet-b7 & MobNet & ViT-B & Swin-B & MaxViT & \\
\midrule
\multirow{4}{*}{RN18}
&AdvLaser   & -&17.88&	32.62	&33.77	&27.81	&24.5	&21.85	&23.51	&25.99
  \\
& AdvShadow    &-& 16.23	&36.09&	39.4	&34.77	&33.77	&27.98	&34.44	&31.81
  \\
& RFLA & -&\textbf{46.36}	&\textbf{70.36}	&\textbf{69.04}	&\textbf{70.2}	&\textbf{63.74}	&24.67	&34.44	&\textbf{54.12}
  \\
& \textbf{AdvWT (Ours)}     &-& 31.29	&57.45	&52.48	&51.49	&48.51	&\textbf{57.78}	&\textbf{59.27}	&51.18
 \\\cmidrule(lr){1-11}
 \multirow{4}{*}{RN50}
&AdvLaser   & 19.37	&-&36.42	&36.92	&35.60	&26.82	&26.66	&21.36	&29.02
  \\
& AdvShadow    & 5.63&-&	35.09	&39.07	&34.77	&31.79&	30.13	&32.78	&29.89
  \\
& RFLA & 28.81	&-&\textbf{64.90}&	\textbf{61.75}&	\textbf{65.56}	&\textbf{54.97}	&25.00&	33.61&	47.80
  \\
& \textbf{AdvWT (Ours)}     & \textbf{37.88}&	-&69.42&	66.54&	69.62&	\textbf{65.96}&	\textbf{69.23}&	\textbf{72.31}&	\textbf{63.59}
 \\\cmidrule(lr){1-11}
\multirow{4}{*}{DN121}
&AdvLaser   & 7.45	&8.94	&-&34.60	&31.62	&22.35	&30.13	&23.84	&22.70
  \\
& AdvShadow    & \textbf{39.19}	&	7.21&-&49.54 &	41.59&	36.97&35.12	&	39.37&35.57
  \\
& RFLA & 9.44	&11.59	&-&\textbf{54.47}	&\textbf{55.96}	&40.23	&23.18	&29.64	&32.07
  \\
& \textbf{AdvWT (Ours)}   &  25.50&	\textbf{28.97}&	-&52.98&	51.32&	\textbf{45.20}&	\textbf{53.97}&	\textbf{55.63}	&\textbf{44.80}
 \\\cmidrule(lr){1-11}
\multirow{4}{*}{EffNet-b7}
&AdvLaser   &34.66	&10.8	&45.64	&-&34.09	&24.62	&37.12	&24.43	&30.19
  \\
& AdvShadow    &\textbf{35.79}	&6.67	&37.37	&-&35.26	&30.53	&32.81	&37.37	&30.83
  \\
& RFLA & 8.11&	13.08	&\textbf{60.76}	&-&\textbf{55.79}	&41.72	&23.84	&31.95	&33.61
  \\
& \textbf{AdvWT (Ours)}   & 24.83	&\textbf{29.30}	&55.96	&-&51.66	&\textbf{45.36}	&\textbf{53.81}	&\textbf{56.46}&	\textbf{45.34}
 \\\cmidrule(lr){1-11}
\multirow{4}{*}{MobNet}
&AdvLaser   & 	\textbf{45.38}&	12.61	&56.72	&46.22	&-&30.67	&42.86	&31.51	&37.99
  \\
& AdvShadow    & 	44.61&	6.99	&48.02	&50.09&-	&37.05	&38.19	&43.67	&38.37
  \\
& RFLA & 7.45	&10.93	&56.79	&50.50&-	&37.91	&22.52	&28.81&	30.70
  \\
& \textbf{AdvWT (Ours)}   & 25.00&	\textbf{28.97}&	\textbf{57.78}	&\textbf{53.64}	&-&\textbf{45.86}	&\textbf{54.80}	&\textbf{57.12}&	\textbf{46.17}
 \\\cmidrule(lr){1-11}
 \multirow{4}{*}{ViT-B}
&AdvLaser   & 13.08	&16.06	&36.26&	32.45	&34.93	&-&29.14	&25.33	&26.75
  \\
& AdvShadow    & 2.98	&8.44	&41.89	&45.70&	42.72&-	&33.61	&37.58	&30.42
  \\
& RFLA & 9.11	&15.89	&53.97	&49.50	&49.83	&-&22.35	&29.30	&32.85
  \\
& \textbf{AdvWT (Ours)}   &  \textbf{27.65}	&\textbf{34.44}	&\textbf{61.26}	&\textbf{58.77}&	\textbf{59.11}&-&	\textbf{60.93}&	\textbf{61.26}&	\textbf{51.92}
 \\\cmidrule(lr){1-11}
\multirow{4}{*}{Swin-B}
&AdvLaser   & \textbf{35.00}	&11.3	&45.00&	35.65&	32.61&	28.70&	-&28.26&	25.98
  \\
& AdvShadow    & 	32.77&	6.33&	35.38&	39.66&	32.96&	33.52&-&	43.95 &	32.08
  \\
& RFLA & 8.11&	11.92	&52.65	&46.52	&47.85&	40.73	&-&30.30&	34.01
  \\
& \textbf{AdvWT (Ours)}   & 25.33	&\textbf{29.14}	&\textbf{56.95}	&\textbf{51.82}	&\textbf{52.65}&	\textbf{45.70}&-&	\textbf{57.78}	&\textbf{45.62}
 \\\cmidrule(lr){1-11}
\multirow{4}{*}{MaxViT}
&AdvLaser   &\textbf{ 40.00}	&11.75&	51.00&	47.50&	40.50&	30.75&	45.00&	-&38.07
  \\
& AdvShadow    & 	34.98	&5.68	&37.73	&46.15	&37.18	&34.07	&38.46	&-&33.46
  \\
& RFLA & 8.94	&11.09	&58.11	&52.65	&\textbf{53.15}	&41.89	&22.68	&-&35.50
  \\
& \textbf{AdvWT (Ours)}   & 24.83&	\textbf{29.64}	&\textbf{56.29}	&\textbf{53.15}	&51.49	&\textbf{45.36}	&\textbf{54.80}&-&	\textbf{45.08}\\
\bottomrule
\end{tabular}
\end{table*}
\begin{table*}[t]
\centering
\setlength{\tabcolsep}{5.2pt}
\caption{Evaluation of \textit{AdvWT} perturbation transferability across various black-box victim models on GTSRB-HQ dataset}\label{tab:table3}
\scriptsize
\begin{tabular}{llcccccccc@{\hspace{6pt}}c}
\toprule
\multirow{2}{*}{Surrogate} & \multirow{2}{*}{Attack} &
\multicolumn{8}{c}{Black-box Target Models} & \multirow{2}{*}{Avg. ASR} \\
\cmidrule(lr){3-10}
 & &RN18 & RN50 & DN121 & EffNet-b7 & MobNet & ViT-B & Swin-B & MaxViT & \\
\midrule
\multirow{4}{*}{RN18}
&AdvLaser   & -&16.18	&16.18	&14.56	&11.65	&7.77	&10.36	&79.06	&12.25
  \\
& AdvShadow    &-&17.48	&21.36	&25.24	&25.57	&21.36	&23.62	&17.15	&21.68
  \\
& RFLA & -&13.74	&\textbf{31.39}	&\textbf{27.51}	&30.09	&\textbf{24.27}	&16.18	&15.53	&22.67
  \\
& \textbf{AdvWT (Ours)}     &-&\textbf{42.30}&	26.89	&26.56	&\textbf{32.79}	&20.33	&\textbf{31.15}&	\textbf{33.11}&	\textbf{30.45}
\\\cmidrule(lr){1-11}
\multirow{4}{*}{RN50}
&AdvLaser   & 29.77	&-&23.95	&26.21	&23.30&	14.24&	20.71	&13.59	&21.68
  \\
& AdvShadow    & 24.27&	-&18.12	&26.21	&22.33&	20.71&	28.80&	20.06&	22.93
  \\
& RFLA & 32.36	&-&\textbf{27.83}	&\textbf{27.51}	&31.39	&\textbf{21.36}	&13.59	&12.30&	23.76
  \\
& \textbf{AdvWT (Ours)}     & \textbf{32.89}	&-&27.24	&25.91	&\textbf{31.56}	&20.60&	\textbf{30.56}&	\textbf{29.57}	&\textbf{28.33}
 \\\cmidrule(lr){1-11}
\multirow{4}{*}{DN121}
&AdvLaser   & 36.25	&32.36&-	&31.39	&27.18	&14.24	&21.68	&17.48	&25.80
  \\
& AdvShadow    & 24.92&	13.74&-&	27.18&	29.13	&22.33&	35.28&	25.89&	25.50
  \\
& RFLA & 16.18&	21.85&-&	\textbf{39.16}&	\textbf{45.63}&	\textbf{24.27}&	20.39&	16.83&	26.33
  \\
& \textbf{AdvWT (Ours)}   &  \textbf{35.62}&	\textbf{47.26}&-&	29.45&	34.59&	21.92&	\textbf{35.96}&	\textbf{35.27}	&\textbf{34.30}
 \\\cmidrule(lr){1-11}
\multirow{4}{*}{EffNet-b7}
&AdvLaser   & 	29.78	&30.42	&22.65&-	&24.60&	21.04&	22.01&	16.50&	23.86
  \\
& AdvShadow    & 20.39	&12.09&	14.89&-&	26.87&	18.12&	30.09&	22.01&	20.64
  \\
& RFLA & 19.09&	17.09	&7.54&	-&\textbf{53.27}	&\textbf{31.16}	&28.14	&26.63&	26.13
  \\
& \textbf{AdvWT (Ours)}   & \textbf{33.33}	&\textbf{37.85}	&\textbf{29.17}&-&	32.64&	22.57&	\textbf{31.60}&	\textbf{30.90}&	\textbf{31.15}
 \\\cmidrule(lr){1-11}
\multirow{4}{*}{MobNet}
&AdvLaser   & 	33.01	&35.92	&29.13	&34.30&-&	14.89	&21.36&	19.09	&26.81
  \\
& AdvShadow    & 23.62	&12.09	&16.83	&24.27&-&	21.04&	30.09&	22.33&	21.47
  \\
& RFLA &            \textbf{43.69}	&18.71	&\textbf{39.48}	&\textbf{35.60}&-&	\textbf{26.54}&	21.04	&18.45&	29.07
  \\
& \textbf{AdvWT (Ours)}   & 37.20&	\textbf{44.37}&	30.72&	28.33&-&	22.18&	\textbf{33.11}&	\textbf{34.47}&	\textbf{32.91}
 \\\cmidrule(lr){1-11}
 \multirow{4}{*}{ViT-B}
&AdvLaser   & 26.21&	28.48	&23.95&	25.57	&21.36	&-&15.21	&16.83	&22.52
  \\
& AdvShadow    & 28.16	&13.25	&16.83	&29.45	&27.83	&-&31.72	&20.39	&23.95
  \\
& RFLA & 33.33	&14.40	&\textbf{35.92}	&32.36	&30.42	&-&19.74	&18.12	&26.33
  \\
& \textbf{AdvWT (Ours)}   &  \textbf{44.61}	&\textbf{49.44}	&35.69	&\textbf{35.69}	&\textbf{32.64}	&-&\textbf{40.89}	&\textbf{46.84}&	\textbf{40.83}
\\\cmidrule(lr){1-11}
\multirow{4}{*}{Swin-B}
&AdvLaser   & 	21.68&	19.42	&17.48	&17.80	&15.86	&11.97&-&	14.89&	17.01
  \\
& AdvShadow    & 	17.15	&10.09	&16.18	&24.92	&18.77	&16.18	&-&22.98	&18.04
  \\
& RFLA & 25.57&	11.09	&24.27	&\textbf{35.89}	&30.74	&21.36	&-&15.21&	23.45
  \\
& \textbf{AdvWT (Ours)}   & \textbf{33.45}&	\textbf{37.59}	&\textbf{27.93}	&30.69	&\textbf{31.38}	&\textbf{24.14}&-&	\textbf{32.76}&	\textbf{31.13}
 \\\cmidrule(lr){1-11}
\multirow{4}{*}{MaxViT}
&AdvLaser   & 18.45	&22.33	&17.80	&19.74	&19.42	&12.30	&19.42	&-&18.49
  \\
& AdvShadow    & 17.80&	13.41	&18.77	&23.30&	23.62&	22.65&	33.66&-&	21.89
  \\
& RFLA & 30.74	&14.24	&26.86	&\textbf{28.80}&	30.74&	\textbf{22.65}&	18.77	&-&24.69
  \\
& \textbf{AdvWT (Ours)}   & \textbf{31.71}	&\textbf{39.37}	&\textbf{28.57}	&27.18	&\textbf{31.01}	&20.91	&\textbf{34.49}	&-&\textbf{30.46}
\\
\bottomrule
\end{tabular}
\end{table*}
\begin{table}[!t]
\centering
\caption{Human study results comparing the perceived naturalness}
\label{tab:table4}
\scriptsize
\begin{tabular}{cc}
\toprule
\multicolumn{1}{c}{\textbf{Method}} & \textbf{Naturalness}\\ \midrule
Real Image                          & 4.17  $\pm$ 1.13             \\ \hdashline
AdvLaser \cite{Duan_2021_CVPR}                           & 2.17  $\pm$ 1.23          \\
AdvShadow \cite{Zhong_2022_CVPR}                           & 1.66    $\pm$ 1.03         \\ 
\midrule
\textbf{AdvWT (Ours)}                       & \textbf{3.66 $\pm$ 1.26}    \\ \bottomrule
\end{tabular}
\end{table}
\begin{figure*}[!t]
    \centering
        \includegraphics[width=0.75\linewidth]{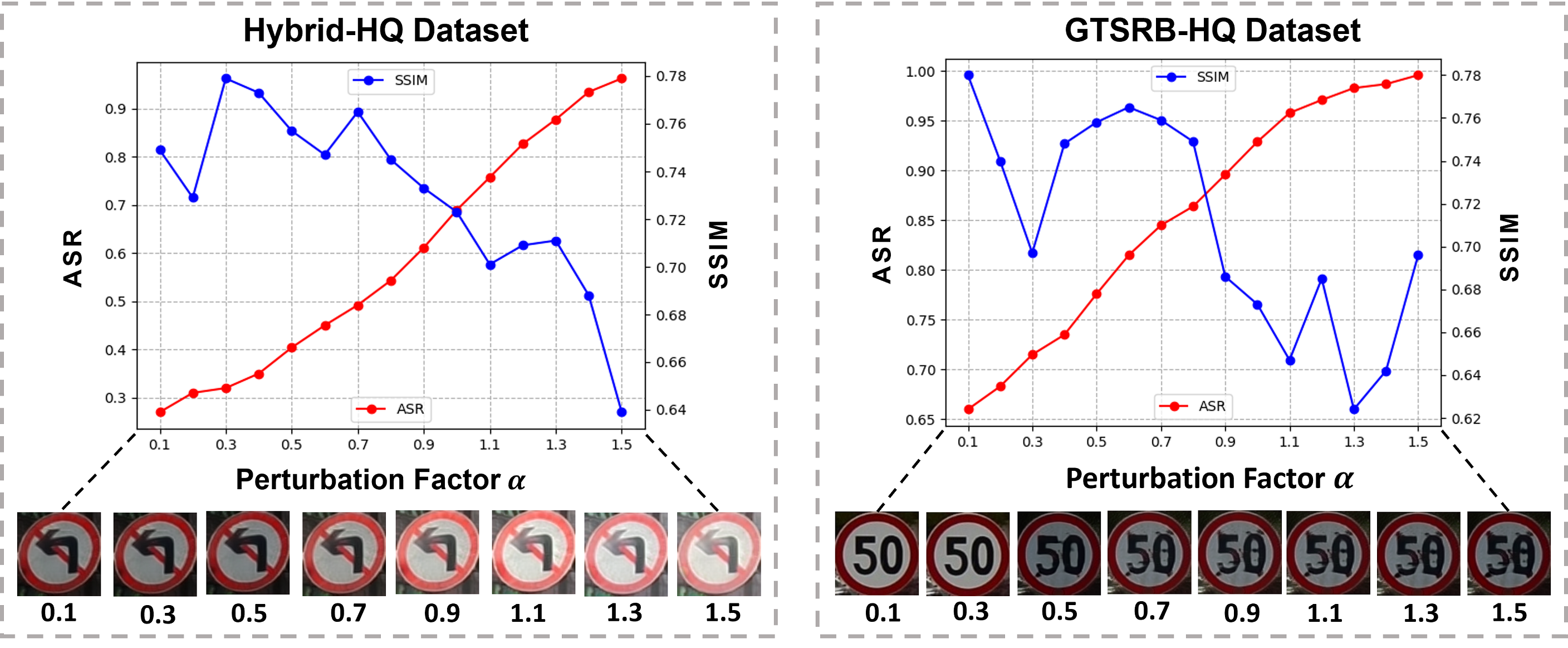}
    \caption{Trade-off between adversarial strength and perceptual similarity: As the perturbation factor $\alpha$ increases, Attack Success Rate (ASR) improves (red line), while structural similarity (SSIM) with the clean image decreases (blue line). This illustrates the inherent balance between damage severity and visual realism in \textit{AdvWT}.}
    \label{fig:fig4}
\end{figure*}
\subsection{Ablation Study of Perturbation Strength $\alpha$}
\textit{AdvWT} controls wear-and-tear severity via a scaling factor 
(\( \alpha \)). Increasing \( \alpha \) improves adversarial success but reduces perceptual fidelity, revealing a trade-off between attack strength and realism. Fig. \ref{fig:fig4} illustrates this trend: ASR (red color) rises steadily with larger perturbations, while SSIM (blue color) falls, reflecting greater visual deviation from clean traffic signs. Instead of a subjective perceptual threshold, we set the maximum perturbation strength (\(\alpha\)) to achieve approximately 98\% attack success on the test set and use SSIM as a proxy for realism. Fig. \ref{fig:fig4} illustrates progressive traffic sign damage as (\( \alpha \)) increases. Additional analysis is provided in Appendix D (Fig. S5) of the Supplementary Materials.
\subsection{Qualitative Results} 
\begin{figure}
    \centering
        \includegraphics[width=0.99\linewidth]{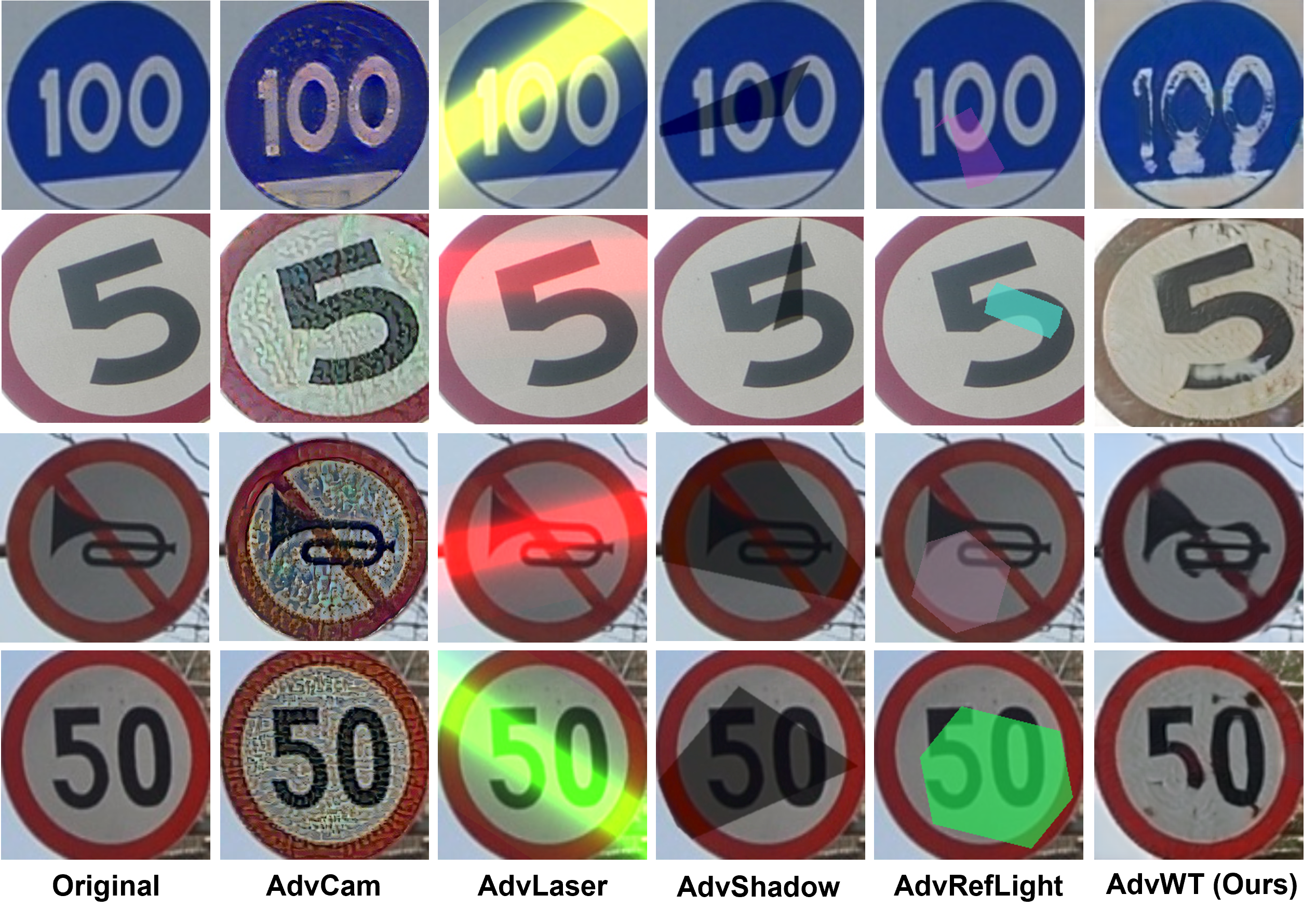}
    \caption{Visualization of adversarial examples generated by \textit{AdvWT} and other methods: \textit{AdvWT} can generate more diverse and realistic adversarial examples since the perturbation is embedded into the sign board through a natural evolving process instead of being crafted by interaction of external projection or light.}
    \label{fig:fig5}
\end{figure}

We qualitatively compare \textit{AdvWT} with other methods by visualizing adversarial examples in Fig. \ref{fig:fig5}. We notice that \textit{AdvWT} generates natural-looking damaged traffic signs that closely resemble real-world wear and tear, making the simulated damage visually indistinguishable from actual deterioration. In comparison, AdvLaser \cite{Duan_2021_CVPR} introduces high-intensity laser patterns that appear visually conspicuous and may partially obscure the sign. AdvShadow \cite{Zhong_2022_CVPR} generates shadow-based occlusions that are effective but often exhibit simplified geometries compared to real-world shadows. AdvCam \cite{Duan2020} applies subtle texture perturbations that can still be distinguishable upon close inspection, while RFLA \cite{Wang2023} produces reflection-based artifacts that may resemble artificial lighting effects on traffic signs.
\subsection{Human Study for Naturalness Evaluation}\label{sec:naturalness_exp}
We conducted a human subjective study with 32 participants to assess the perceived naturalness of \textit{AdvWT} images.\\
\textbf{Participant details:} Participants were aged 23–35 (mean: 28.3); 24 identified as male and 8 as female. Seventeen reported a background in computer vision, while the remainder did not.\\
\textbf{User study protocol and Results:}
Image order and method identity were randomized and blinded. Images were displayed at 256$\times$256 resolution with no time limit. Each participant rated 10 image groups, each containing a real traffic sign and adversarial examples generated by AdvLaser, AdvShadow, and \textit{AdvWT} using a 5-point Likert scale (1 = highly unrealistic, 5 = highly realistic). Each image received 32 ratings. Table \ref{tab:table4} reports the mean and standard deviation across participants. \textit{AdvWT} achieves high naturalness scores, closely matching the perceived realism of real traffic signs.\\
\textbf{Statistical Analysis:} To assess whether the methods received statistically different ratings, we employed the Friedman test, a non-parametric test for multiple paired comparisons. The test revealed a significant effect of method on naturalness ratings ($\chi^2(3)=72.33$, $p=1.35\times10^{-15}$), indicating systematic differences in median ranks across methods. The extremely small $p$-value confirms that these differences are unlikely to be due to chance, indicating that participants consistently perceived differences in naturalness across methods.
\subsection{Physical Domain Experiments}\label{sec:physical_exp}
\begin{figure}
    \centering
        \includegraphics[width=0.9\linewidth]{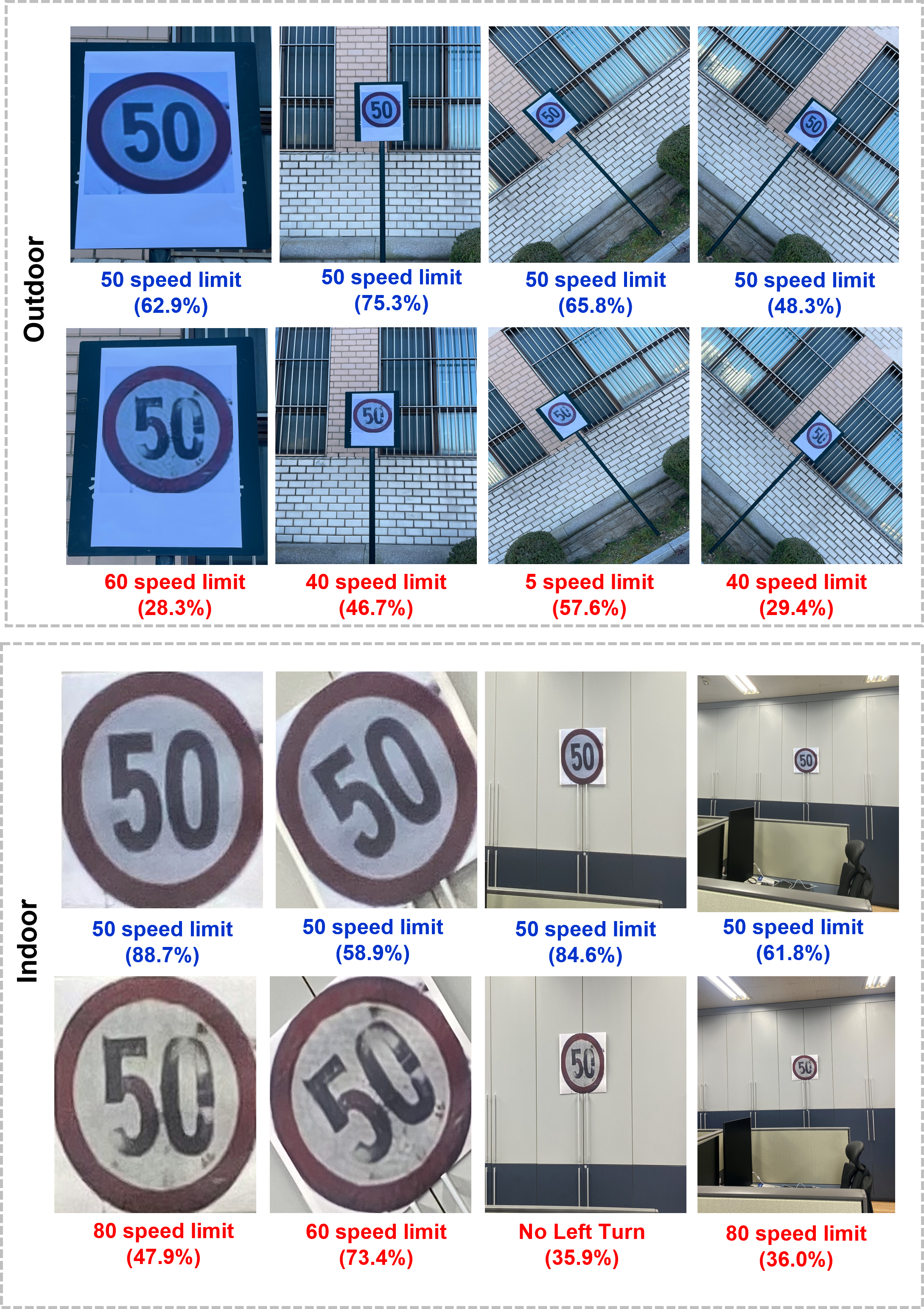}
    \caption{Physical-world evaluation of \textit{AdvWT} perturbation: We attack `50 speed limit' sign and evaluate under indoor and outdoor environments. We show a clean sign (top row) and a corresponding adversarial sign (bottom row). Each sign is shown with the predicted label and its corresponding confidence score. Correct predictions are shown in blue font, while incorrect predictions are indicated in red font.}
    \label{fig:fig6}
\end{figure}

To evaluate real-world applicability, we test whether \textit{AdvWT} perturbations remain effective beyond the digital domain under indoor and outdoor conditions. Clean and adversarial `50 speed limit' signs are printed using an EPSON L8180 printer and photographed with an iPhone 11 from varying distances and angles, after which the captured images are analyzed by a trained traffic sign classifier. The results of our physical-world experiments are illustrated in Fig. \ref{fig:fig6}. Each image is labeled with its predicted class and the classifier's confidence score. As shown, \textit{AdvWT} generates realistic degradations that are sufficient to mislead the classifier under varying physical conditions, including different distances and viewpoints. Furthermore, \textit{AdvWT} adversarial examples remain potent even after real-world transformations such as printing and recapturing.
\section{Further Analysis}
\subsection{Defending against \textit{AdvWT}}\label{sec:defense_exp}
In the preceding section, we demonstrated that \textit{Wear and Tear} perturbations severely compromise traffic sign recognition. We now assess whether \textit{AdvWT} can be mitigated by different classes of defenses. We evaluate \textit{AdvWT} against Adversarial Training (AT) \cite{Madry2017}, and alternative purification-based defenses, including Diffusion-based Adversarial Purification (Diff-AP) \cite{nie2022diffusion}, and localized implicit function-based purification (DISCO) \cite{ho2022disco}. AT improves robustness by incorporating adversarial examples during training. Diff-AP removes perturbations through iterative denoising with a pretrained diffusion model, while DISCO performs deterministic local reconstruction via implicit functions. To train the model robustly, we apply \textit{AdvWT}-based modifications to 50\% of training images and retrain ResNet-18 models, denoted as Hybrid\(_{rob}\) and GTSRB-HQ\(_{rob}\). As shown in Table \ref{tab:table5}, AT substantially reduces vulnerability to \textit{AdvWT}, yet the attack success rate remains around 50\% on the Hybrid dataset. Notably, adversarially trained models also improve accuracy on clean data, suggesting that \textit{AdvWT} serves as an effective augmentation strategy for real-world robustness. To apply Diff-AP \cite{nie2022diffusion}, we encode clean and adversarial images into the latent space of a diffusion model \cite{rombach2022high}, perform latent diffusion-based denoising, and decode the purified latents back into images. Image-conditioned prompts are generated using BLIP-2 \cite{li2023blip}. DISCO \cite{ho2022disco} is trained on paired clean and corresponding \textit{AdvWT} images. Table \ref{tab:table6} reports the standard and robust accuracy of purification-based defenses on \textit{AdvWT}, where standard accuracy is measured on purified clean samples and robust accuracy on purified adversarial samples. Both Diff-AP and DISCO significantly improve robustness against \textit{AdvWT} compared to no defense. While current defense mechanisms can enhance robustness, there remains substantial room for improvement against advanced, naturalistic adversarial attacks such as \textit{AdvWT}.
\begin{table}[!t]
\centering
\scriptsize
\caption{Robustness of \textit{AdvWT} under Adversarial Training}
\label{tab:table5}
\begin{tabular}{lcc}
\toprule
\textbf{Method} & \multicolumn{2}{c}{\textbf{Accuracy (\%)}}  \\ 
\cmidrule(lr){2-3}
\multicolumn{1}{c}{}                        & \textbf{Benign}        & \textbf{\textit{AdvWT-AT} }  \\ \midrule
Hybrid                               & 96.5          & 4.6            \\
Hybrid$_{rob}$                       & \textbf{98.0} & \textbf{50.7} \\ \midrule
GTSRB-HQ                                & 99.6          & 1.3 
           \\
GTSRB-HQ$_{rob}$                       & \textbf{99.6}          & \textbf{36.4}        \\ 
\bottomrule
\end{tabular}
\end{table}

\begin{table}[!t]
\centering
\scriptsize
\caption{Robustness of \textit{AdvWT} under Adversarial Purification}
\label{tab:table6}
\begin{tabular}{lccc}
\toprule
\textbf{Dataset} & \textbf{Method} & \multicolumn{2}{c}{\textbf{Accuracy (\%)}}  \\ 
\cmidrule(lr){3-4}
\multicolumn{1}{c}{}  & \multicolumn{1}{c}{}                       & \textbf{Standard Acc.}        & \textbf{Robust Acc.}  \\ \midrule
\multirow{3}{*}{Hybrid} & No defense   &    \textbf{96.2}    &     3.48     \\
&      Diff-AP                & 91.72 &  \textbf{52.48}\\ 
&      DISCO                & \textbf95.69 & 39.07 \\ 
\midrule
\multirow{3}{*}{GTSRB} & No defense   &    \textbf{99.68}    &     0.33     \\
&      Diff-AP                & 86.41 & \textbf{48.54} \\ 
&      DISCO                & 89.96 & 47.93 \\ 
\bottomrule
\end{tabular}
\end{table}
\subsection{Attack Efficiency}\label{sec:attack_eff}
\begin{figure}
\centering
\includegraphics[width=0.99\linewidth]{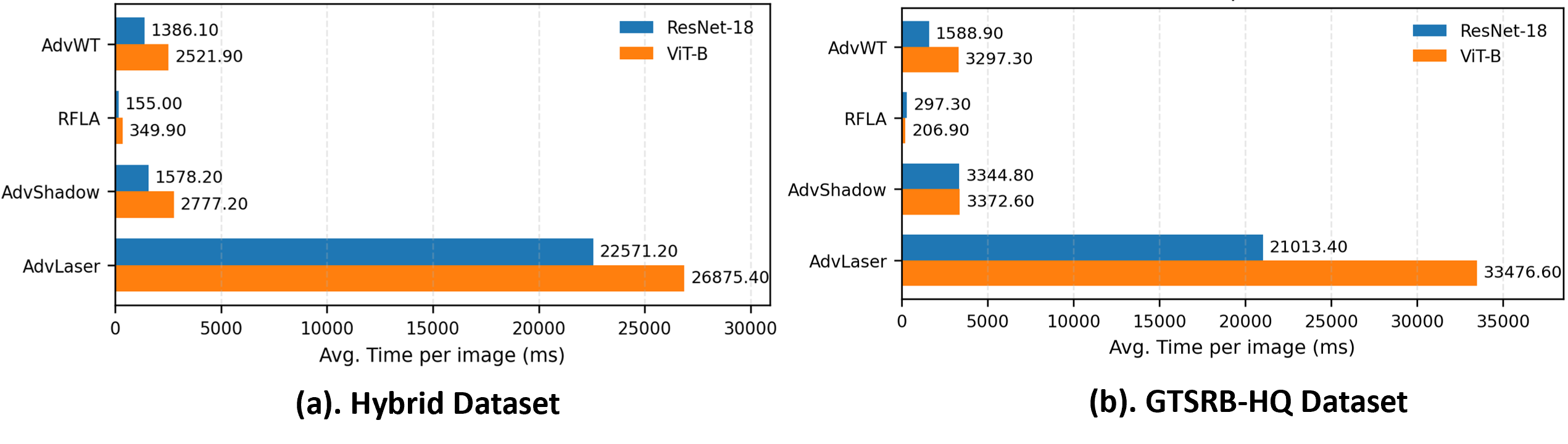}
\caption{Comparison of average per-image attack time (ms) on two target models: ResNet-18 (blue) vs. ViT-B (orange). Values on bars are means. Lower is better.}\label{fig:fig7}
\end{figure}

We evaluate the efficiency of \textit{AdvWT} by measuring the wall-clock time required to generate adversarial examples and comparing it with prior attacks \cite{Duan_2021_CVPR,Zhong_2022_CVPR,Wang2023}. Fig. \ref{fig:fig7} reports the mean attack time per image across two datasets (Hybrid, GTSRB-HQ) and two architectures (ResNet-18, ViT-B). \textit{AdvWT} exhibits moderate runtime, slower than RFLA but significantly faster than AdvLaser and AdvShadow. Attacks on ViT-B are consistently slower than on ResNet-18, reflecting the higher computational cost of transformer models. While RFLA is fastest due to lightweight pixel-space perturbations, \textit{AdvWT} incurs additional cost from latent-space search and image synthesis.
\subsection{Out-of-Distribution Generalization}\label{sec:ood_exp}
Real-world damaged traffic signs pose challenges for DNNs in safety-critical applications. To evaluate whether training with \textit{AdvWT}-generated damage improves Out-of-Distribution (OOD) generalization, we conduct experiments on real-world damaged traffic signs and compare three models:
\begin{enumerate}
\item \textbf{Baseline:} Trained without augmentation.  
\item \textbf{Baseline + Corrupt:} Trained with simulated common visual corruptions \cite{Hendrycks2019}, where each corruption type has five levels of severity. We introduce corruptions randomly per image, selecting both the corruption type and severity level.  
\item \textbf{Baseluine + \textit{AdvWT}:} Trained using \textit{AdvWT}-generated damaged signs as augmentation.  
\end{enumerate}
For both common corruptions and \textit{AdvWT} augmentations, perturbations are applied with 50\% probability. As shown in Table \ref{tab:table7}, training with \textit{AdvWT} significantly improves generalization to real-world damaged traffic signs compared to baseline and corruption-based augmentation, highlighting the effectiveness of semantically meaningful adversarial damage. 
\begin{table}[!t]
\centering
\caption{OOD generalization on real-world damaged traffic signs}
\scriptsize
\label{tab:table7}
\begin{tabular}{lc}
\toprule
\multicolumn{1}{c}{\textbf{Model}} & \multicolumn{1}{c}{\textbf{Accuracy (\%})} \\ \midrule
Baseline                      & 90.1                         \\
Baseline + Corrupt              & 93.4                         \\
Baseline + \textit{AdvWT}             & \textbf{95.4}                         \\ \bottomrule
\end{tabular}
\end{table}
\begin{figure}
    \centering
        \includegraphics[width=0.8\linewidth]{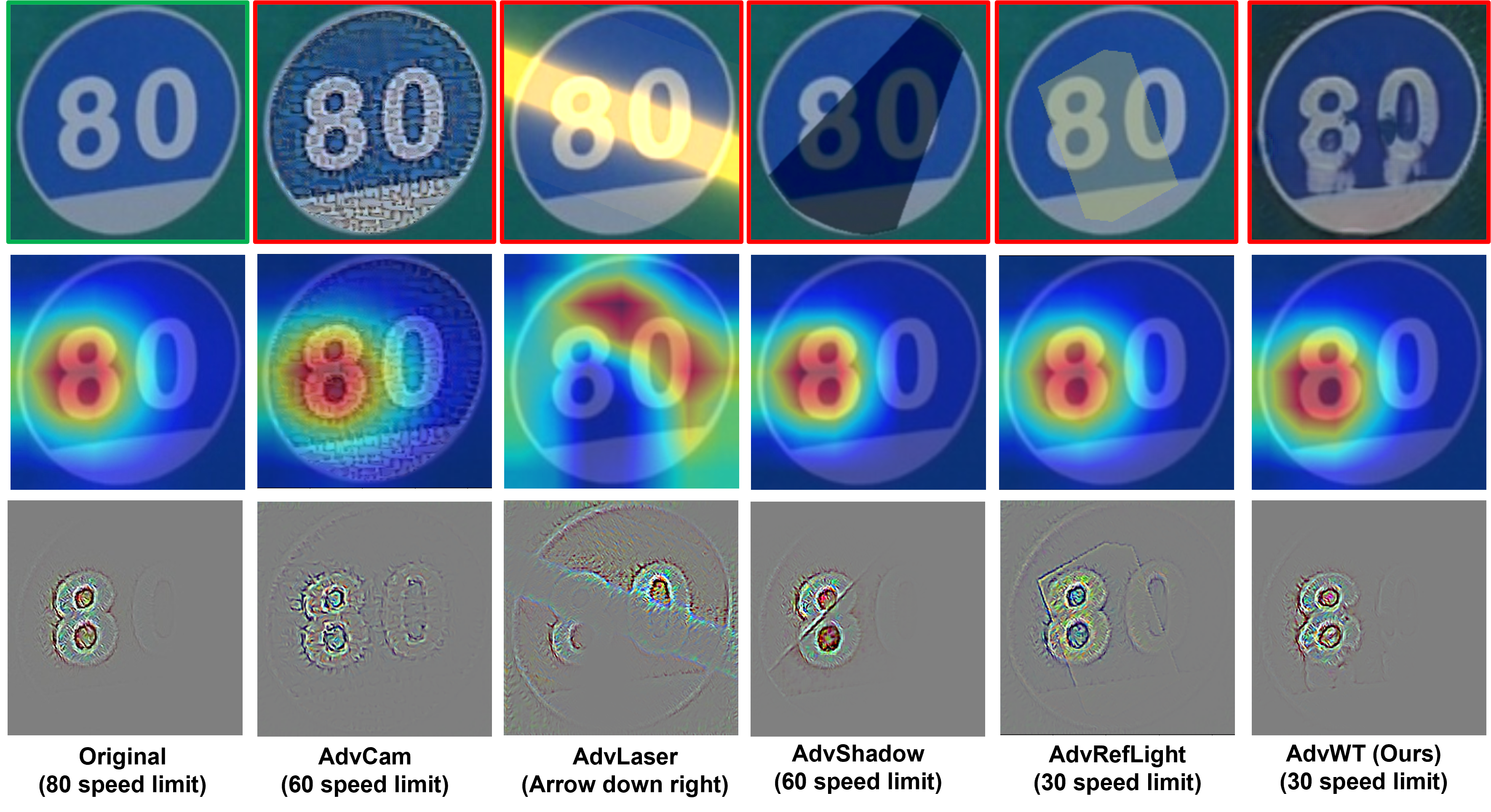}
    \caption{Interpretability analysis using Grad-CAM and Guided Grad-CAM: Visualizing how different adversarial attacks influence the predictions of road sign classifier.}
    \label{fig:fig8}
\end{figure}
\begin{figure}
\centering
\includegraphics[width=0.8\linewidth]{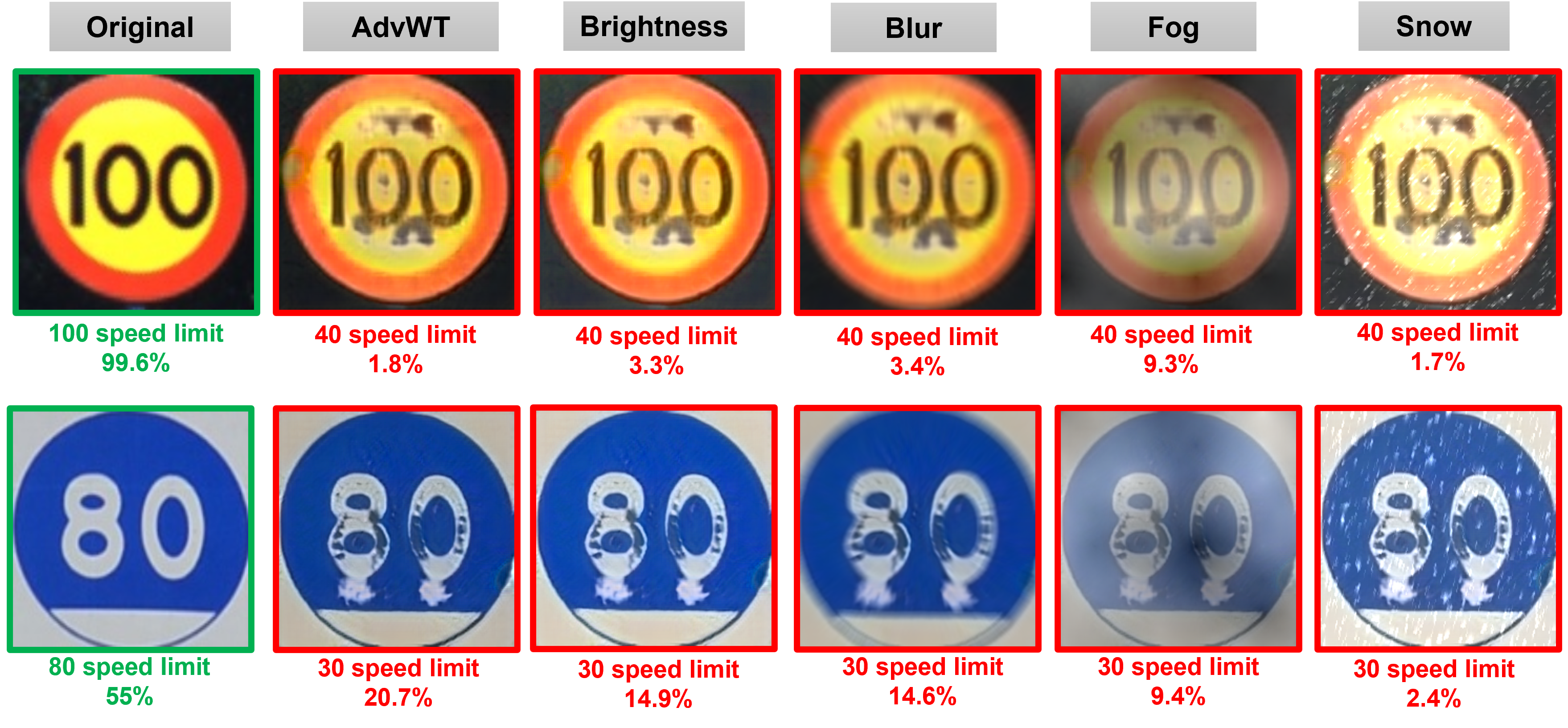}
\caption{Visual examples showing the persistence of \textit{AdvWT} under common environmental corruptions (brightness, blur, fog, snow) for \textbf{80} and \textbf{100} speed limit signs.}\label{fig:fig9}
\end{figure}
\begin{figure}[!t]
    \centering
        \includegraphics[width=0.8\linewidth]{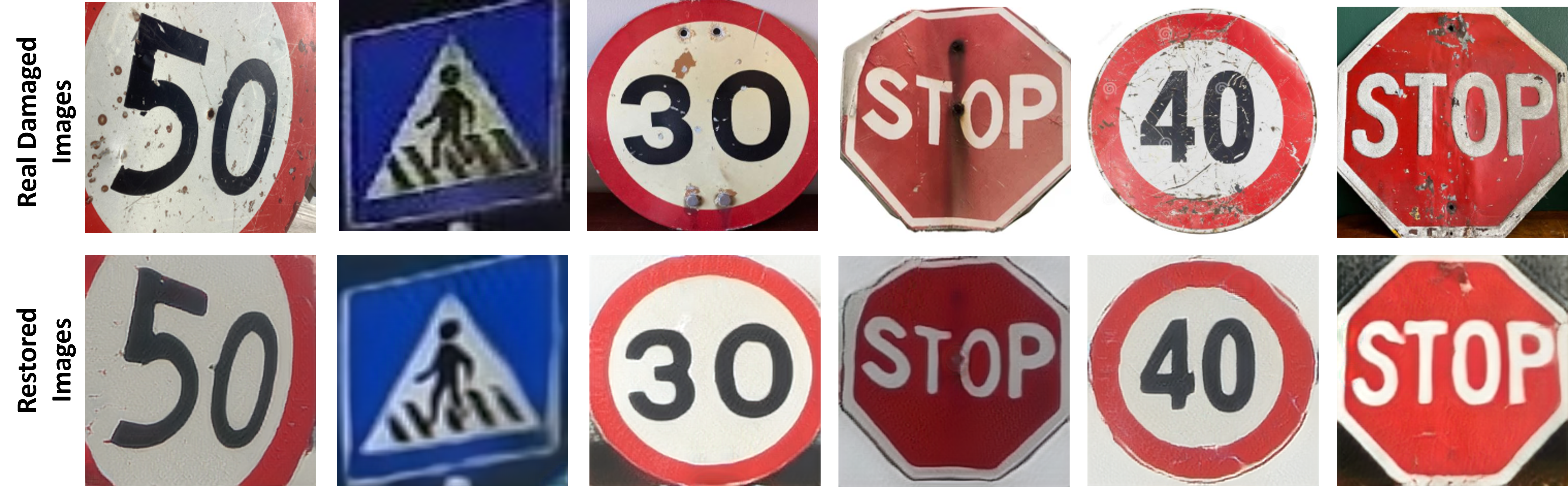}
    \caption{Traffic sign restoration by reversing real-World damage}
    \label{fig:fig10}
\end{figure}
\subsection{Interpreting Predictions of Model}\label{sec:gradcam_exp}
We use Grad-CAM and Guided Grad-CAM to analyze how different attacks influence classifier attention \cite{Selvaraju2016}. As shown in Fig. \ref{fig:fig8}, the benign sign is correctly classified with attention focused on the digit `8'. AdvCam \cite{Duan2020} induces subtle attention spread, while AdvLaser \cite{Duan_2021_CVPR} shifts attention to irrelevant regions due to strong distortions. AdvShadow \cite{Zhong_2022_CVPR} largely preserves relevant focus but introduces misleading cues at shadow boundaries, and RFLA \cite{Wang2023} highlights edges of synthetic reflections. In contrast, \textit{AdvWT} maintains attention on semantically important regions while still causing misclassification.
\subsection{Stability across Weather and Visual Transformations}\label{sec:stable_exp}
To evaluate real-world applicability, we test the robustness of \textit{AdvWT} under common environmental corruptions, including brightness shifts, blur, fog, and snow \cite{Hendrycks2019}. These transformations simulate realistic outdoor conditions and are applied to adversarial traffic sign images. Fig. \ref{fig:fig9} shows results for the `\textbf{80}' and `\textbf{100}' speed limit signs. \textit{AdvWT} perturbations remain effective under all corruptions, with the classifier consistently misclassifying the images despite minor confidence fluctuations (e.g., a drop from 20\% to 14\% under fog). This demonstrates that \textit{AdvWT} perturbations persist beyond clean conditions and remain robust to environmental distortions.
\subsection{Analysis of Damageness of traffic signs}\label{sec:damageness}
We define natural damage as environmentally induced wear from factors such as sunlight, rain, and dust, resulting in irregular fading, cracks, corrosion, and dirt, in contrast to externally imposed perturbations like lasers, shadows, or patches. We characterize natural damage using the empirical distribution of real damaged signs and assess realism via FID. As a baseline, we use Natural Color Fool (NCF) \cite{yuan2022natural}, which alters color without structural wear, and additionally evaluate on an externally damaged sign set. As shown in Table \ref{tab:table8}, \textit{AdvWT} achieves substantially lower FID than NCF with non-overlapping 95\% bootstrap confidence intervals on both Internal (25.46 vs. 37.01) and External (71.64 vs. 89.34) datasets, indicating closer alignment with real-world damage in- and out-of-domain. We further assess adversarial image quality using FID \cite{heusel2017gans}, LPIPS \cite{zhang2018unreasonable}, and SSIM \cite{ssim} (Table \ref{tab:table9}). Both datasets yield realistic adversarial signs: Hybrid shows slightly better perceptual and structural fidelity (LPIPS/SSIM), while GTSRB-HQ achieves marginally lower FID.
\begin{table}[t]
\centering
\caption{Comparison of distribution similarity (CLIP-FID) between real and generated damaged traffic signs, reported with 95\% CIs.}
\scriptsize
\begin{tabular}{lcc}
\toprule
Method & CLIP-FID $\downarrow$ & 95\% CI \\
\midrule
NCF vs. Real-damaged (Internal)   & 37.01 & [37.34, 39.57] \\
\textbf{AdvWT vs. Real-damaged (Internal)} & \textbf{25.46} & \textbf{[26.27, 27.89]} \\
\midrule
NCF vs. Real-damaged (External)  & 89.34 & [90.59, 95.32] \\
\textbf{AdvWT vs. Real-damaged (External)} & \textbf{71.64} & \textbf{[74.03, 77.97]} \\
\bottomrule
\end{tabular}
\label{tab:table8}
\end{table}
\begin{table}[t]
\centering
\caption{Image quality evaluation using FID (lower is better), LPIPS (lower is better), and SSIM (higher is better)} \label{tab:table9}
\setlength{\tabcolsep}{8pt}
\scriptsize
\begin{tabular}{lccc}
\toprule
\textbf{Dataset} & \textbf{FID} $\downarrow$ & \textbf{LPIPS} $\downarrow$ & \textbf{SSIM} $\uparrow$ \\
\midrule
Hybrid & 80.01 & 0.2104 & 0.778 \\
GTSRB-HQ & 75.23 & 0.2247 & 0.751 \\
\bottomrule
\end{tabular}
\end{table}
\subsection{Structural Analysis of AdvWT perturbations}\label{sec:advwt_pert}
Our method models naturalistic wear-and-tear effects (e.g., cracks, fading, and dirt), producing non-uniform perturbations that may enhance stealthiness and cross-model transferability. To understand the underlying reasons for these improvements, we analyze the optimized adversarial perturbations themselves, i.e., the injected noise used to fool the model, in both the frequency and spatial domains. In the frequency domain, we compute frequency entropy from the Fourier spectrum of the perturbation to assess energy dispersion \cite{zhang2021universal} \cite{yin2019fourier}. Specifically, we compute the perturbation $\delta=x_{adv} - x_{clean}$, apply a 2D Fourier transform, aggregate spectral energy into orientation bins, and compute the Shannon entropy of the resulting distribution \cite{zhang2021universal}. In the spatial domain, we measure perturbation coverage, the ratio of the largest perturbed region, and elongation to characterize the structure and distribution of adversarial artifacts \cite{wang2021interpreting}. Perturbation coverage quantifies the spatial extent of the perturbation and is computed as the fraction of pixels whose perturbation magnitude exceeds a fixed threshold. The largest perturbed region ratio measures the degree of spatial concentration and is computed as the area of the largest connected perturbed component divided by the total perturbed area. Elongation quantifies the geometric shape of perturbations and is computed as the ratio of the major to the minor eigenvalue of the covariance matrix for each connected perturbed region. 

Table \ref{tab:table10} reports Fourier entropy, Spatial Coverage, Largest-region Ratio, and Elongation. \textit{AdvWT} achieves higher frequency entropy than template-driven attacks, including AdvLaser \cite{Duan_2021_CVPR}, AdvShadow \cite{Zhong_2022_CVPR}, and RFLA \cite{Wang2023}, indicating more spectrally dispersed perturbations. This observation is consistent with prior Fourier-based analyses of adversarial perturbations, which suggest that spectrally distributed perturbations tend to reflect broader, model-agnostic vulnerabilities rather than narrow, model-specific artifacts \cite{zhang2021universal} \cite{yin2019fourier}. Spatially, \textit{AdvWT} shows higher coverage with lower dominance and elongation, suggesting less localized and less structured perturbations. Overall, these results demonstrate that \textit{AdvWT} produces more broadly distributed and natural perturbation patterns, providing quantitative evidence of improved transferability and stealth.
\begin{table}[t]
\centering
\caption{Analysis of perturbation patterns in the frequency and spatial domains. \textbf{Entropy} corresponds to Fourier entropy, \textbf{Spat. Cov.} to spatial coverage, \textbf{Large Ratio} to the largest-region ratio, and \textbf{Elong.} to elongation.} \label{tab:table10}
\setlength{\tabcolsep}{6pt}
\scriptsize
\begin{tabular}{@{}lcccc@{}}
\toprule
 \textbf{Attack} & \textbf{Entropy} $\uparrow$ & \textbf{Spat. Cov.} $\uparrow$ & \textbf{Large Ratio} $\downarrow$ & \textbf{Elong.} $\downarrow$ \\
\midrule
AdvLaser&   4.534            &  26.610        & 0.682 & 2.433\\
AdvShadow & 10.389           & 27.02           & 0.517 & 1.846 \\
RFLA&       14.215           & \textbf{49.279} & 0.460 & 2.047\\
AdvWT (Ours)&      \textbf{16.499}   & 35.10          & \textbf{0.284} & \textbf{1.787}\\
\bottomrule
\end{tabular}
\end{table}
\subsection{Application to Traffic Sign Restoration}
Since our GAN model learns bidirectional mappings between undamaged and damaged domains, we also evaluate its potential for traffic sign restoration. By applying undamaged-domain style codes to real-world damaged signs, the model generates restored images that resemble their original state. As shown in Figure \ref{fig:fig10}, the model effectively removes natural wear and tear, suggesting practical applicability beyond adversarial attacks, including automated traffic sign restoration.
\subsection{Discussion and Future Work}
The Adversarial Wear and Tear (\textit{AdvWT}) attack introduces a novel class of physical adversarial perturbations that mimic real-world degradation patterns such as fading, corrosion, and scratches to deceive DNNs. Unlike prior methods that apply arbitrary style or latent manipulations, \textit{AdvWT} is the first to model real-world, safety-relevant degradation patterns, creating structured and physically grounded adversarial perturbations. However, the proposed framework models degradation patterns assuming single-material objects, such as traffic signs. As a result, it cannot yet simulate complex wear patterns arising from multi-material interactions, such as objects composed of glass, metal, and plastic, where degradation manifests differently across materials. Additionally, the diversity of real-world wear-and-tear datasets remains limited. Although \textit{AdvWT} is trained to produce physically plausible degradation patterns, the availability of large-scale, labeled datasets capturing real-world damage (e.g., corrosion, erosion, or long-term exposure effects) is scarce. This constraint may limit the representational diversity of the generated perturbations. Furthermore, \textit{AdvWT} focuses on visual appearance degradation and does not model other physical factors such as geometric deformation, illumination-dependent material changes, or temporal evolution of damage, which may further influence real-world perception systems. Finally, as \textit{AdvWT} relies on GAN-based degradation modeling, there is a risk of overfitting to the degradation patterns. While our experiments demonstrate transferability across models and datasets, this remains an inherent limitation of learned generative approaches. \\
\textbf{Future Direction:} While this work focuses on adversarially damaged traffic signs, physical degradation naturally manifests in diverse ways across objects. For instance, smartphones with fractured screens, faded printed documents, animals covered in mud or cobwebs, rotten fruit, or human faces obscured by dust in industrial settings. Our current design is specialized for traffic signs due to their safety-critical role in autonomous driving, but the underlying principle of modeling semantically plausible degradations is broadly applicable. Future research will extend this framework to other domains, further exploring naturally occurring adversarial vulnerabilities and their impact on deep learning systems.
\section{Conclusion}
In this work, we introduce \textit{AdvWT}, the first approach to simulate real-world physical damage on traffic signs for generating adversarial examples. Unlike conventional physical-world adversarial examples that rely on ad-hoc perturbations, \textit{AdvWT} leverages the natural `wear and tear' process as an adversarial pattern. Due to the `continuous evolving' characteristic of the physical damage, we formulate the problem of `damaged traffic sign generation' as a `damage representation learning' problem. We successfully simulate the realistic damages on the traffic signs by leveraging unsupervised GAN-based image-to-image translation model and demonstrate that structured physical degradation can effectively fool DNNs. Our experiments in both digital and physical settings confirm the attack effectiveness, transferrability, stealthiness, and robustness of \textit{AdvWT}, highlighting its potential as a realistic and naturally occurring adversarial threat. Additionally, we demonstrate that training with \textit{AdvWT}-augmented data enhances a model's generalizability to real-world damaged traffic signs, presenting a novel avenue for improving robustness in safety-critical applications.

\section{Biography Section}
\begin{IEEEbiography}[{\includegraphics[width=1.23in,height=1.23in,clip,keepaspectratio]{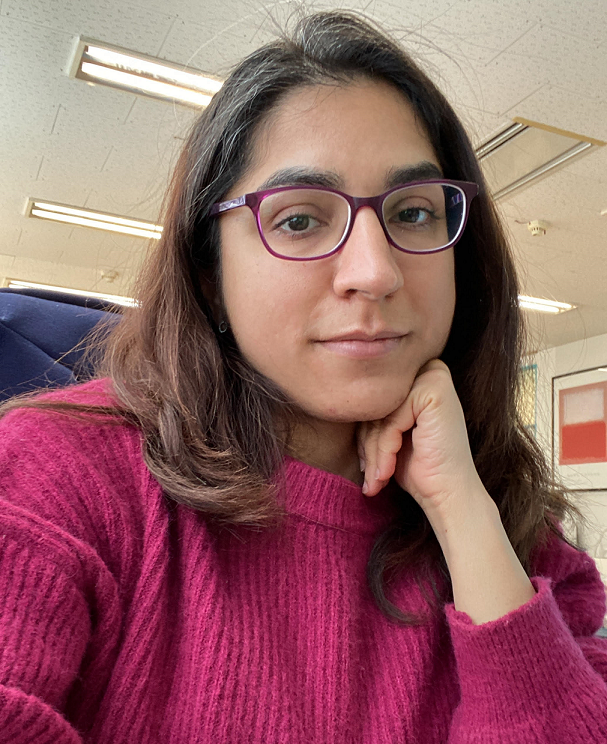}}]{Samra Irshad} received a B.S. degree in 2012 from COMSATS University, Lahore, Pakistan, and an M.S. in 2015 from the National University of Sciences and Technology, Islamabad, Pakistan. She served as a Lecturer in the Computer Science and Engineering department, at Superior University, Lahore, from 2015-2016. From 2017-2018, she was with Victoria University, Melbourne, Australia, and researched evolutionary algorithms and ensemble learning with an application in retinal imaging. She was with Swinburne University of Technology, Melbourne, Australia, from 2018-2020 and researched 3D medical image segmentation. From 2021-2022, she worked as a machine learning engineer in a medical AI startup in Melbourne, Australia. She is currently pursuing her Ph.D. research in the Augmented Intelligence Lab at Kyung Hee University, Yongin-si, South Korea. Her research interests include generative modelling, AI safety, and robust deep learning.
\end{IEEEbiography} 
\begin{IEEEbiography}[{\includegraphics[width=1in,height=1.25in,clip,keepaspectratio]{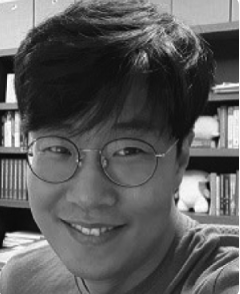}}]{Seungkyu Lee} received the B.S. and M.S. degrees in electrical engineering from the Korea Advanced Institute of Science and Technology, Daejeon, South Korea, in 1997 and 1999, respectively, and the Ph.D. degree in computer science and engineering from The Pennsylvania State University, in 2009. He has been a Principal Research Scientist with the Advanced Media Laboratory, Samsung Advanced Institute of Technology, Yongin, South Korea. He is currently a Professor with Kyung Hee University. His research interests include generative neural network, color/depth image processing, computer vision and machine learning and 3-D reconstruction.
\end{IEEEbiography} 
\begin{IEEEbiography}
[{\includegraphics[width=1in,height=1.25in,clip,keepaspectratio]{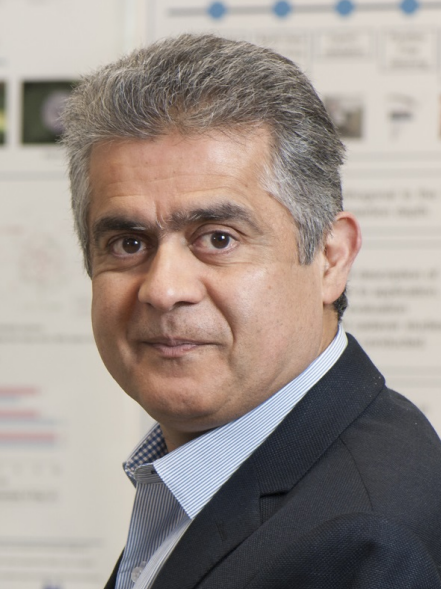}}]{Nassir Navab} is a Full Professor and Director of the Laboratory for Computer-Aided Medical Procedures at Johns Hopkins University and the Technical University of Munich. He has also secondary faculty appointments at both affiliated Medical Schools. He completed his PhD at INRIA and University of Paris XI, France, and enjoyed two years of a post-doctoral fellowship at MIT Media Laboratory before joining Siemens Corporate Research (SCR) in 1994. At SCR, he was a distinguished member and received the Siemens Inventor of the Year Award in 2001. He received the SMIT Society Technology award in 2010 and the ‘10 years lasting impact award’ of IEEE ISMAR in 2015. In 2012, he was elected as a Fellow of the MICCAI Society. He has acted as a member of the board of directors of the MICCAI Society, 2007-2012 and 2014-2017, and serves on the Steering committee of the IEEE Symposium on Mixed and Augmented Reality (ISMAR) and Information Processing in Computer-Assisted Interventions (IPCAI). In 2021, he was elected as a Fellow of the IEEE Society. He is the author of hundreds of peer-reviewed scientific papers, with more than 88,000 citations and an h-index of 127.
\end{IEEEbiography}
\begin{IEEEbiography}
[{\includegraphics[width=1in,height=1.25in,clip,keepaspectratio]{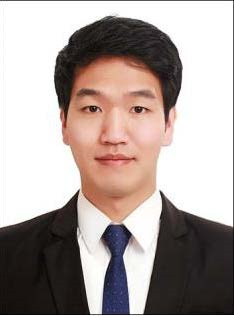}}]{Hong Joo Lee} is an Assistant Professor at the Seoul National University of Science and Technology, South Korea. He received the B.S. degree from Ajou University, Suwon, South Korea, in 2016, and the M.S. and Ph.D. degrees from the Korea Advanced Institute of Science and Technology (KAIST), Daejeon, South Korea, in 2018 and 2023, respectively. He was a postdoctoral researcher at the Technical University of Munich (TUM), Munich, Germany, from 2023~2025. His research interests include deep learning, machine learning, medical image segmentation, and adversarial robustness.
\end{IEEEbiography}
\begin{IEEEbiography}
[{\includegraphics[width=1in,height=1.25in,clip,keepaspectratio]{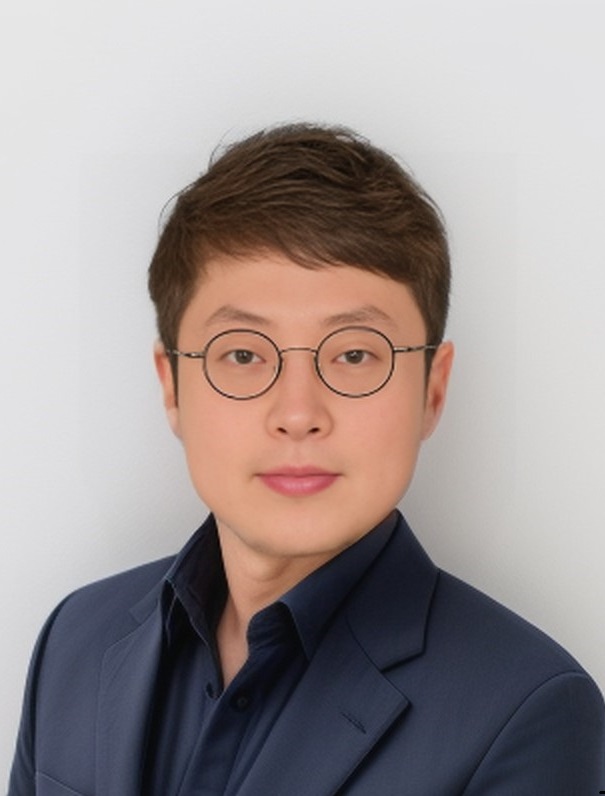}}]{Seong Tae Kim} is an Assistant Professor in the Department of Computer Science and Engineering, Kyung Hee University, Yongin-si, South Korea. He received B.S. degree from Korea University, Seoul, South Korea, in 2012, and the M.S. and Ph.D. degree from the Korea Advanced Institute of Science and Technology (KAIST), Daejeon, South Korea, in 2014 and 2019, respectively. In 2015, he was a visiting researcher with the University of Toronto, ON, Canada. From 2019 to 2021, he was a Senior Research Scientist in the Chair for Computer Aided Medical Procedures, Technical University of Munich, Munich, Germany. His current research interests include multimodal AI, explainable AI, and data-efficient deep learning. He authored or coauthored more than 60 peer-reviewed journal and conference papers.
\end{IEEEbiography}

\vfill

\end{document}